\documentclass{article}

\PassOptionsToPackage{authoryear,round}{natbib}

\usepackage[preprint]{neurips_2026}

\usepackage[utf8]{inputenc}
\usepackage[T1]{fontenc}

\usepackage{amsmath}
\usepackage{amssymb}
\usepackage{amsfonts}
\usepackage{amsthm}
\usepackage{amstext}
\usepackage{mathtools}
\usepackage{bm}
\usepackage{dsfont}
\usepackage{bbm}
\usepackage{nicefrac}

\usepackage{graphicx}
\usepackage{booktabs}
\usepackage{multirow}
\usepackage{longtable}
\usepackage{array}
\usepackage{threeparttable}
\usepackage{float}
\usepackage{wrapfig}
\usepackage{caption}
\usepackage{subcaption}    \usepackage{pgfplots}
\pgfplotsset{compat=1.18}

\usepackage{algorithm}
\usepackage[noend]{algpseudocode}

\usepackage{verbatim}
\usepackage{comment}
\usepackage[inline]{enumitem}
\usepackage{autobreak}
\usepackage{microtype}
\usepackage{xcolor}        \usepackage{hyperref}
\usepackage{cleveref}
\usepackage{url}

\usepackage[subtle,tracking=normal,wordspacing=normal,mathspacing=normal,mathdisplays=normal]{savetrees}

\newcommand{\E}{\mathbb{E}}

\newcommand{\Pjoint}{P_{XY}}
\newcommand{\Pmarg}{P_X \otimes P_Y}
\newcommand{\KL}{\mathrm{KL}}
\newcommand{\TV}{\mathrm{TV}}

\newcommand{\PMI}{\mathrm{PMI}}

\newcommand{\fMI}{I_f}

\newcommand{\strategy}{\sigma}

\newtheorem*{theorem*}{Theorem}
\newtheorem{theorem}{Theorem}[section]

\newtheorem{example}[theorem]{Example}
\newtheorem{proposition}[theorem]{Proposition}
\newtheorem{definition}[theorem]{Definition}

\title{Scoring Rules! Statistical and Strategic Alignment \\
for Text Evaluation Metrics}

\author{Shengwei Xu\thanks{Both authors contributed equally to this paper.}\\
University of Michigan\\
\texttt{shengwei@umich.edu}\\
\And
Yuxuan Lu$^*$\\
Peking University\\
\texttt{yx\_lu@pku.edu.cn}\\
\And
Yifan Wu\\
Microsoft Research\\
\texttt{yifan.wu2357@gmail.com}\\
\And
Jason Hartline\\
Northwestern University\\
\texttt{hartline@eecs.northwestern.edu}\\
\And
Grant Schoenebeck\\
University of Michigan\\
\texttt{schoeneb@umich.edu}\\
}

\begin{document}

\maketitle

\begin{abstract}
Reference-based text evaluation metrics, which are widely used to assess natural language generation systems, score a candidate response by comparing it with a reference response. The reliability of an evaluation metric is usually judged by the statistical correlation with human ratings. However, as these metrics are increasingly used as optimization objectives, this alignment by correlation is no longer sufficient: agents may strategize to game the evaluation metric. We study this issue through two complementary notions of alignment. A metric is \emph{statistically aligned} if it correlates with human ratings, and \emph{strategically aligned} if it resists strategic perturbations that do not add task-relevant information. 

We make two contributions. First, we propose test principles for reference-based metrics, consisting of human-rating correlation, degradation sensitivity, and manipulation robustness. These principles evaluate whether a metric agrees with human judgments, penalizes low-effort information loss, and resists strategic score inflation. Second, we develop a unified design framework for mutual-information-based metrics that are designed to resist strategic perturbations. The framework decomposes existing and new metrics into four design choices: information measure, estimation method, text representation, and prediction mechanism. 
Across peer review, summarization, and question answering tasks, we find that strong human-rating correlation does not imply strategic alignment: LLM-as-a-Judge achieves high correlation but is susceptible to manipulations. In contrast, mutual-information-based metrics 
substantially improve manipulation robustness. Our framework also uncovers a new 
metric that achieves the strongest overall robustness in our experiments while remaining competitive on human-rating correlation.
\end{abstract}

\section{Introduction}

Reference-based evaluation metrics are essential for scalable assessment of natural language generation (NLG) systems. Given a candidate response and one or more reference responses for the same task, such metrics assign a score intended to reflect the quality, informativeness, or correctness of the candidate. Human judgement of text quality remains the gold standard, but it is too slow and expensive at large scale. 
Automatic evaluation metrics therefore serve as scalable proxies for human judgment: they replace repeated human assessment with a computable comparison between the candidate and reference responses. Examples include lexical-overlap metrics such as BLEU \citep{papineni2002bleu} and ROUGE \citep{lin2004rouge}, embedding-based metrics such as BERTScore \citep{zhang2019bertscore}, and LLM judges that compare a candidate response against a reference. 
Throughout, an \emph{evaluation metric} is a function that assigns a numerical score to a candidate using one or more references. We use \emph{evaluation workflow} for the datasets, perturbations, and statistical tests used to assess metrics.
Most standard reference-based metrics are designed to approximate human judgments. We call this requirement \textbf{statistical alignment}:  a useful metric should correlate with human assessments of text quality.

However, reference-based metrics create incentives for language generation systems, which may not align with the intended target of improving genuine text quality. As their role expands from measurement to optimization, we need to understand the incentives these scores create. Prior work has documented related failure modes under the names of reward hacking \citep{amodei2016concrete}, specification gaming \citep{krakovna2020specification}, or reward overoptimization \citep{gao2023scaling}, where optimizing a proxy objective can produce behavior that increases the measured score without improving the intended target.

We call the complementary requirement \textbf{strategic alignment}: a metric should be robust to strategic perturbations of the text. In particular, a metric should assign lower expected scores to \emph{less informative} reports, which omit or degrade task-relevant content, and to \emph{untruthful} reports, which manipulate the evaluation procedure without genuinely improving the answer. Recent work proposes evaluation metrics robust to strategic manipulations \citep{lu2024eliciting, xu2025benchmarking, robertson2025beyondvibes, wu2024elicitationgpt}. These works hint that evaluation metrics must be judged not only by the statistical correlation with human ratings, but also by whether they are robust to strategic perturbations of the text.

\paragraph{Contribution 1: Test Principles} Existing metrics for manipulation robustness are often developed and evaluated in isolation, using different datasets, perturbations, and success criteria. As a result, direct comparison is difficult. 
It remains unclear how evaluation metrics should be systematically validated when the goal is not only statistical alignment with human ratings, but also strategic alignment under score-seeking behavior. 

In this paper, we address this validation gap by introducing unified test principles for reference-based evaluation metrics. Generalized from previous studies \citep{lu2024eliciting, xu2025benchmarking, robertson2025beyondvibes, wu2024elicitationgpt}, we propose three complementary principles for comparing metrics. The first principle is \emph{correlation} with human preference: a metric should agree with human judgments under ordinary, non-adversarial conditions. This principle corresponds to \text{statistical alignment} and captures the standard desideratum in natural-language generation (NLG) evaluation.

The second and third principles evaluate \text{strategic alignment} from two complementary perspectives. The second principle is \emph{degradation sensitivity}: when a text is deliberately degraded by removing or corrupting task-relevant information, the metric score should strictly decrease. This tests whether the metric discourages \emph{less informative} reports, which save effort by providing strictly less information while attempting to receive similar credit. 

The third principle is \emph{manipulation robustness}: when a text is strategically modified to inflate the metric score without adding genuine content, the metric should not increase. This tests whether the metric discourages \emph{untruthful} reports, which exploit superficial features of the evaluation procedure. Unlike degradation, manipulation need not strictly reduce the information content of the text. It may only weakly decrease it or preserve much of the original content while changing style, format, or wording to obtain a higher score. Together, these three principles provide a common standard for evaluating both prior and new metrics on equal footing. 

We instantiate these principles in a reusable evaluation workflow over seven datasets spanning peer review, summarization, and question answering. Four datasets include human quality ratings for correlation tests, and all seven support degradation and manipulation tests.

\paragraph{Contribution 2: Metric Design Framework} Having introduced a unified test framework, we now turn to the complementary construction question: how should one design metrics that pass these tests? At a high level, \textit{mutual information} (MI) measures how much information two variables share; we give the formal definition in \Cref{sec:preliminaries}. 
MI is a natural target because it is known to be information-monotone: transformations that do not add task-relevant information do not increase the information a candidate shares with an independent reference. MI-based metrics use this idea by scoring a candidate through an estimate of shared information with the reference.
Existing methods such as GEM \citep{xu2025benchmarking}, GPPM \citep{lu2024eliciting}, and TVD-MI \citep{robertson2025beyondvibes} can be read as different approximations to this common target. We defer the technical distinctions among them to the design framework; the point here is that they vary along a shared set of design choices.

However, there is no unified framework for constructing strategically aligned metrics. Previous papers are developed as separate methods rather than as instances of a common design principle.
For example, GEM \citep{xu2025benchmarking} uses the token representation of the text, GPPM-Judgment \citep{lu2024eliciting} parses the text as a set of statements, while TVD-MI \citep{robertson2025beyondvibes} sends the whole piece of text into an LLM-oracle to estimate mutual information.   Thus, they interpret the response as represented by different sized chunks---token, statement, response, respectively.  
Additionally, they differ in the choice of the MI measure, estimator, and prediction model. 
Because prior evaluations typically compare the full methods rather than controlled variations of these individual components, it is difficult to determine which design choices are responsible for the observed behavior.

We propose a unified design framework for constructing MI-based evaluation metrics.  Our framework decomposes every MI-based metric into four components: an information measure (MI), an MI estimation method, a representation of a text, and a prediction mechanism. This decomposition recovers existing methods such as GEM, GPPM, and TVD-MI as special cases, while also exposing a broader design space of new metrics.
The framework therefore gives a principled recipe for constructing new metrics and for analyzing which design components lead to desirable properties.

Our empirical results support a central takeaway: statistical alignment alone is not enough to certify an evaluation metric as strategically aligned. 
Across peer review, summarization, and question answering, we find that LLM-as-a-Judge achieves the strongest correlation with human ratings, yet is vulnerable to manipulations. In contrast, MI-based metrics provide stronger strategic alignment. In particular, the best-performing new metric from our design framework computes MI between candidate and reference texts at the statement level. It is more reliable than existing MI-based metrics such as GEM, GPPM, and TVD-MI. 
Therefore, our framework is constructive, not only descriptive: it uncovers a new MI metric that outperforms prior MI-based metrics in our evaluation workflow.

Together, our two contributions provide a unified perspective on aligned text evaluation beyond statistical alignment with human judgements. The test principles explain how evaluation metrics should be validated, while the design framework explains how MI-based metrics can be constructed and varied. Combining the two allows us to compare existing methods fairly, explore new points in the design space, and identify which combinations of information measure, estimator, and distribution model lead to the most reliable evaluation behavior across domains.

\section{Preliminaries}\label{sec:preliminaries}

Let \(\mathcal{V}\) be a vocabulary, and let \(\mathcal{X}=\mathcal{Y}=\mathcal{V}^{*}\) denote the spaces of candidate and reference texts. Let \(\mathcal{W}\) denote the task space. Throughout, uppercase letters denote random variables and lowercase letters denote their realizations. Thus, \(W\) is a task random variable with task distribution \(P_W\), while \(w\in\mathcal{W}\) denotes a realized task.

Conditional on the realized task \(W=w\), the candidate random variable \(X\), supported on the candidate-text space \(\mathcal{X}\), and the reference random variable \(Y\), supported on the reference-text space \(\mathcal{Y}\), are drawn according to
\[
X\sim P_{X\mid W=w},
\qquad
Y\sim P_{Y\mid W=w}.
\]
We assume that the candidate $X$ and reference $Y$ are conditionally independent given the task $W$.
The distributions \(P_X\), \(P_Y\), and \(P_{XY}\) denote the candidate marginal distribution, reference marginal distribution, and joint candidate--reference distribution induced by this generative process.

A \textit{reference-based evaluation metric} is a function
\[
S:\mathcal{X}\times\mathcal{Y}\times
\bigl(\mathcal{Y}\cup\{\emptyset\}\bigr)
\longrightarrow \mathbb{R}.
\]
For a realized candidate text \(x\in\mathcal{X}\), matched reference text \(y\in\mathcal{Y}\), and optional negative reference \(y^-\in\mathcal{Y}\cup\{\emptyset\}\), the scalar \(s=S(x,y,y^-)\) is the resulting metric score. We write \(S(x,y)\) as shorthand for the noncontrastive score \(S(x,y,\emptyset)\).

For contrastive metrics, the negative-reference random variable \(Y^-\sim P_Y\) is drawn from the marginal reference distribution \(P_Y\), independently of the task, candidate, and matched-reference random variables \((W,X,Y)\). Thus, the realized negative reference \(y^-\) represents a reference from an independently sampled task. For noncontrastive metrics, we set \(y^-=\emptyset\).

The negative reference allows \(S\) to reward agreement with the matched reference \(y\) relative to agreement with the unrelated reference \(y^-\). In particular, contrastive scoring discounts textual patterns that are common under the marginal reference distribution \(P_Y\) but are not specific to the realized task \(w\).

The metric-evaluation problem is to assess whether \(S\) satisfies statistical alignment and the two parts of strategic alignment. Statistical alignment asks whether scores on unperturbed candidate responses correlate with human quality judgments. Strategic alignment asks whether scores decrease under degradations that remove task-relevant information and do not increase under manipulations that seek higher scores without adding such information. The rest of the paper develops tests for these properties and a design framework for constructing metrics that target them.

\subsection{Mutual Information and the Data-Processing Inequality}

The guiding principle for our framework is information monotonicity: a metric should not assign a higher expected score after a perturbation that removes or fails to add task-relevant information. Degradations and manipulations are both perturbations of informativeness, which we formally introduce in \Cref{sec:eval-pipeline}. Degradations remove task-relevant content, while manipulations seek a higher score through transformations that do not add task-relevant content.

Under the conditional-independence assumption, the candidate random variable \(X\) and reference random variable \(Y\) share dependence through the task random variable \(W\). Mutual information quantifies this dependence. A candidate response that preserves task-relevant facts, reasoning, or semantic content should therefore share more information with the matched reference than a candidate response from which such information has been removed. We formalize this idea using $f$-mutual information.

\begin{definition}[$f$-divergence and $f$-mutual information]\label{def:f-mi}
Let \(U\) and \(V\) be random variables taking values in a finite or countable sample space \(\Omega\), with respective distributions \(P_U\) and \(P_V\) and probability mass functions \(p_U\) and \(p_V\). Assume that \(\operatorname{supp}(P_U)\subseteq\operatorname{supp}(P_V)\); equivalently, \(p_V(z)=0\) implies \(p_U(z)=0\). For a convex generator \(f:\mathbb{R}_{+}\to\mathbb{R}\) satisfying \(f(1)=0\), the $f$-divergence from \(P_V\) to \(P_U\) is
\[
D_f(P_U\|P_V) \coloneqq \sum_{z\in\Omega:\,p_V(z)>0} p_V(z)\, f\left(\frac{p_U(z)}{p_V(z)}\right).
\]

Let \(P_{XY}\) denote the joint distribution of the candidate and reference random variables \((X,Y)\), and let \(P_X\otimes P_Y\) denote the product of their marginal distributions. Their $f$-mutual information is
\[
\fMI(X;Y) \coloneqq D_f\bigl(P_{XY}\,\|\,P_X\otimes P_Y\bigr).
\]
The support condition above holds automatically for this pair of distributions: \(p_{XY}(x,y)>0\) implies \(p_X(x)>0\) and \(p_Y(y)>0\).
\end{definition}

\noindent
Thus, the $f$-mutual information \(\fMI(X;Y)\) measures the divergence between the joint candidate--reference distribution \(P_{XY}\) and the product-of-marginals distribution \(P_X\otimes P_Y\). For the KL-divergence generator \(f(t)=t\log t\), $f$-mutual information recovers Shannon mutual information \citep{cover1999elements}.

\begin{proposition}[Data-processing inequality]\label{prop:dpi-fmi}
Let the transformed-candidate random variable \(X'\) be obtained from the candidate random variable \(X\) through a possibly randomized transformation that has no additional access to the reference random variable \(Y\). Equivalently, suppose that \(Y-X-X'\) forms a Markov chain. Then
\[
\fMI(X';Y)\le \fMI(X;Y).
\]
In particular, for any deterministic candidate transformation \(\sigma:\mathcal{X}\to\mathcal{X}'\),
\[
\fMI(\sigma(X);Y)\le \fMI(X;Y).
\]
\end{proposition}

The data-processing inequality states that post-processing the candidate random variable \(X\) cannot increase its $f$-mutual information with the reference random variable \(Y\). Consequently, a candidate transformation \(\sigma\) that has no access to additional task information can at most preserve, and may reduce, the task-relevant information shared with the reference.

Let \(p_{XY}\) denote the joint probability mass function of the candidate and reference random variables, and let \(p_X\) and \(p_Y\) denote their marginal probability mass functions. The joint-to-product density ratio is the function \(r:\mathcal{X}\times\mathcal{Y}\to\mathbb{R}_{+}\) defined by
\[
r(x,y)
\coloneqq
\frac{p_{XY}(x,y)}{p_X(x)\,p_Y(y)}
=
\frac{p_{Y\mid X}(y\mid x)}{p_Y(y)}
=
\frac{p_{X\mid Y}(x\mid y)}{p_X(x)}
\]
for candidate--reference pairs \((x,y)\) in the support of the product-of-marginals distribution \(P_X\otimes P_Y\). The population $f$-mutual information can be written in terms of the joint-to-product density ratio \(r\) as
\begin{equation}
\label{eq: mi density ratio}
    \fMI(X;Y)
    = \E_{(X,Y)\sim P_X\otimes P_Y}\bigl[f(r(X,Y))\bigr].
\end{equation}

\section{Test Principles}\label{sec:eval-pipeline}

We compare evaluation metrics along statistical alignment and strategic alignment, under our test principles, generalized from previous work \citep{lu2024eliciting,xu2025benchmarking,robertson2025beyondvibes, wu2024elicitationgpt}. Specifically, for statistical alignment, we test the statistical correlation with human quality ratings; for strategic alignment, we use degradation strategies to simulate low-effort reporting and manipulation strategies to simulate untruthful reporting. We test whether the metrics can penalize degradations and resist manipulations.

\paragraph{Statistical alignment test: Correlation with human ratings.} 

For datasets providing absolute human scores, let $h_i$ denote the human rating on candidate response $x_i$, and $s_i$ denote the corresponding score assigned by an evaluation metric. We measure statistical alignment using Spearman's rank correlation,
\[
\rho = \mathrm{Spearman}\big(\{s_i\}_{i=1}^n,\{h_i\}_{i=1}^n\big).
\]
When multiple annotators are available, we average their scores at the item level before computing $\rho$.

\paragraph{Strategic alignment test 1: Sensitivity to degradation}\label{sec:deg-tests}

A \emph{degradation} is a perturbation that reduces the task-relevant information, coverage, or specificity of a candidate by deleting evidence, restricting context, or forcing low-effort responses; a well-behaved metric should assign strictly lower scores to degraded variants.

For each candidate report $x_i$, we apply a degradation strategy $M$ to produce a perturbed report $x_i^{M}$, and compute scores $s_i$ and $s_i^{M}$ for the original and perturbed report respectively. Table~\ref{tab:degradations} lists six strategies used in our experiments in Section~\ref{sec:factorial}.

For each strategy $M$, we use the paired t-test to test whether the mean score change is significantly negative: 
\[
\bar{\Delta}^{(M)} \coloneqq \frac{1}{n}\sum_{i=1}^n s_i^{M} - s_i < 0
\]
To quantify the magnitude of the sensitivity to degradation across rules that use different scales, we also report standardized mean difference (SMD) \citep{andrade2020mean} with the 95\% confidence interval (which shares the same form of Cohen's $d$):
\[
d_M = \frac{\mu_M - \mu}{\sqrt{(\sigma_M^2 + \sigma^2)/2}},
\]
where $(\mu,\sigma)$ and $(\mu_M,\sigma_M)$ are the empirical mean and standard deviation of the original and perturbed scores respectively.

\paragraph{Strategic alignment test 2: Manipulation Resistance}\label{sec:manip-tests}

A \emph{manipulation} is a perturbation designed to inflate the score by exploiting metric shortcuts, such as verbosity, stylistic regularity, or generic content likely under the reference distribution, without introducing new task-specific evidence; a robust metric should not assign higher scores to such variants.

As with the degradation tests, for each candidate report $x_i$, we apply a manipulation strategy $M$ to produce a perturbed report $x_i^{M}$. Table~\ref{tab:manipulations} lists six strategies used in our experiments in Section~\ref{sec:factorial}.

We compute the paired mean score change $\bar\Delta^{(M)}$ and standardized mean difference $d_M$ using the same formulas established for degradations. For manipulations, we test whether the mean score change is significantly positive. A metric passes a manipulation test when the score does not significantly increase after the manipulation, i.e., when we do not find evidence that \(\bar{\Delta}^{(M)} > 0\).

\begin{table}[htbp]
\centering
\footnotesize
\caption{Degradation strategies.}
\label{tab:degradations}
\renewcommand{\arraystretch}{1.18}
\begin{tabular}{@{}p{0.2\linewidth}p{0.55\linewidth}p{0.2\linewidth}@{}}
\toprule
\textbf{Strategy} & \textbf{Description} & \textbf{Provenance} \\
\midrule
Random Replacement & Replace the original candidate with a response sampled from a different task within the same task family. & \citet{lu2024eliciting} \\
Sentence Deletion & Delete every other sentence while preserving section headers, ordering, and visible format. & \citet{xu2025benchmarking} \\
Deletion \& Completion & Delete every other sentence, mark each gap, then ask a helper LLM to fill the gaps using only the remaining text and the original format. & \citet{xu2025benchmarking} \\
Surface Report & Regenerate the response using only weak context, such as the abstract for peer review or the headline plus lead paragraph for summarization. & \citet{robertson2025beyondvibes}, \citet{xu2025benchmarking} \\
Ultra-Concise Compression & Summarize under a severe length cap (e.g., 10\% of the original), so that specific evidence and caveats must be dropped. & \citet{robertson2025beyondvibes} \\
Opinion Flip & Reverse the overall evaluative verdict (e.g., accept $\to$ reject, positive $\to$ negative) while preserving the surface structure and topical vocabulary of the original candidate. & \footnotesize\citet{robertson2025beyondvibes} \\
\bottomrule
\end{tabular}
\end{table}

\begin{table}[htbp]
\centering
\footnotesize
\caption{Manipulation strategies.}
\label{tab:manipulations}
\renewcommand{\arraystretch}{1.18}
\begin{tabular}{@{}p{0.2\linewidth}p{0.55\linewidth}p{0.2\linewidth}@{}}
\toprule
\textbf{Strategy} & \textbf{Description} & \textbf{Provenance} \\
\midrule
Rephrase & Ask an LLM to rewrite the candidate in a specific style while preserving meaning and high-level structure. & \citet{xu2025benchmarking} \\
Meaningless Elongation & Append the same fixed, semantically empty filler sentences to every candidate report without adding task-relevant information. & \citet{xu2025benchmarking} \\
Opinion Shift (Positive) & Rewrite the candidate to preserve structure and coverage while systematically shifting stance toward more favorable sentiment. & \citet{robertson2025beyondvibes}\\
Opinion Shift (Negative) & Rewrite the candidate to preserve structure and coverage while systematically shifting stance toward less favorable sentiment. & \citet{robertson2025beyondvibes}\\
Opinion Shift (Neutral) & Rewrite claims as hedged, noncommittal statements (e.g., ``the method is effective'' $\to$ ``the method may have some merit''), flattening the evaluative signal without changing topical content. & This work \\
Opinion Shift (Extreme) & Amplify all claims to their strongest possible form (e.g., ``a reasonable approach'' $\to$ ``a groundbreaking contribution''), testing whether a metric rewards confidence or extremity as a proxy for quality. & This work \\
\bottomrule
\end{tabular}
\end{table}

\section{A Design Framework for MI-Based Metrics}

The test principles of \Cref{sec:eval-pipeline} specify what an aligned metric should do; this section addresses how to construct one.
We introduce a framework for MI-Based Metrics that separates the design space into four components, divided into two conceptual parts. 
The first part consists of \emph{theoretical design choices}: which information measure is being estimated, and which estimator is used.
The second part consists of \emph{empirical design choices}: how reports are represented, and how predictive quantities are obtained from those representations. 

\begin{definition}[Design tuple]\label{def:triple}
An $f$-MI evaluation metric is specified by a tuple
\[
\mathcal{D}\coloneqq (F,E,R,\Pi),
\]
where:
\begin{enumerate}[label=(\roman*),nosep]

    \item \textbf{Information measure} $F$: the choice of $f$-divergence defining the objective $\fMI(X;Y)$.

    \item \textbf{Estimation method} $E$: the method used to estimate the chosen objective $\fMI(X;Y)$.

    \item \textbf{Representation} $R$: a map from the text space to a representation space,
    \(
    R:\mathcal{V}^{*}\to \mathcal{U}.
    \)

    \item \textbf{Prediction} $\Pi$: a mechanism that maps represented candidate-reference pairs to predictive quantities used in the estimation method. 
\end{enumerate}
\end{definition}

\paragraph{How the components compose.}
The four components form a scoring pipeline with two dependencies. On the theoretical side, the information-measure component $F$ fixes the population objective, and the estimation-method component $E$ prescribes which predictive quantity suffices to estimate it: direct estimation requires the density-ratio function $r(x,y)$, whereas variational estimation requires only the critic function $T^*(x,y)$ (\Cref{subsec:mi-estimation}). On the empirical side, the representation component $R$ fixes the units of text passed to the prediction mechanism $\Pi$, and the prediction mechanism $\Pi$ supplies the quantity that the estimation method $E$ requires. Concretely, to score a realized item $(x_i,y_i,y_i^-)$, the metric (i) applies the representation map $R$ to the texts, (ii) queries the prediction mechanism $\Pi$ on the represented pairs to obtain the predictive quantities, and (iii) plugs those quantities into the estimator $E$ to produce the sample-level score $S_{F,E,R,\Pi}(x_i,y_i,y_i^-)$.
 
\paragraph{Sample-level scores.}
Given a dataset of $n$ tasks $\{w_i\}_{i=1}^n$, each task $w_i$ has an associated candidate report $x_i$ and a positive reference report $y_i$. Designs that score contrastively additionally use a negative reference $y_i^-$ sampled from the marginal distribution $P_Y$, whose agreement with the candidate is penalized; designs that do not use one take $y_i^-=\emptyset$. The sample-level score is obtained from the chosen estimator by replacing the population expectations with the sampled references: the expectation over the joint distribution $\Pjoint$ is evaluated at $(x_i,y_i)$, and the expectation over the product distribution $\Pmarg$ at $(x_i,y_i^-)$. Averaging the per-item scores across the dataset then yields an estimate of the population objective determined by the tuple:
\[
\frac{1}{n}\sum_{i=1}^{n} S_{F,E,R,\Pi}(x_i,y_i,y_i^-) \;\approx\; \fMI(X;Y).
\]

\paragraph{Two running examples.}
Prior MI-based metrics correspond to specific choices of the tuple. To make the pipeline concrete, we trace the components and scoring pipelines of two prior metrics that sit at opposite corners of the design space, and we return to them throughout the section.
 
\begin{example}[GEM: $(\KL,\ \text{Token},\ \text{Autoregression},\ \text{Direct})$]\label{ex:gem}
GEM~\citep{xu2025benchmarking} instantiates the four components as follows.
\(F=\KL\) selects KL mutual information as the population objective. 
\(E=\text{Direct}\) forms the plug-in pointwise mutual information
\[
S(x_i,y_i)=\log p_\phi(y_i\mid x_i)-\log p_\phi(y_i).
\]
\(R=\text{Token}\) represents the candidate report $x_i$ and reference report $y_i$ as token sequences.
\(\Pi=\text{Autoregression}\) uses a language model with parameters $\phi$ to compute the conditional likelihood $p_\phi(y_i\mid x_i)$ token by token with the candidate in the prompt, and the marginal likelihood $p_\phi(y_i)$ with the candidate removed.

Because direct KL estimation takes an expectation only under the joint candidate--reference distribution, no negative reference is needed; the negative-reference input is $y_i^-=\emptyset$.
\end{example}
 
\begin{example}[TVD-MI: $(\TV,\ \text{Full-report},\ \text{LLM-oracle},\ f\text{-variational})$]\label{ex:tvdmi}
TVD-MI~\citep{robertson2025beyondvibes} makes a different choice on every axis.
\(F=\TV\) selects total-variation mutual information as the population objective.
\(E=f\text{-variational}\) contrasts the two judgments to produce the sample-level score
\[
S(x_i,y_i,y_i^-)=T(x_i,y_i)-T(x_i,y_i^-),
\]
\(R=\text{Full-report}\) keeps each report whole.
\(\Pi=\text{LLM-oracle}\) prompts an LLM, e.g., ``Do these two reports describe the same item?'', once on the matched candidate--reference pair $(x_i,y_i)$ and once on the mismatched candidate--reference pair $(x_i,y_i^-)$; its yes/no answer implements a critic function $T$. For total variation, this critic only needs to decide whether the density-ratio value $r(x,y)$ exceeds $1$.

This metric rewards candidates that the oracle can link to their own task's reference but not to an unrelated one. Because the variational form takes expectations under both the joint and the product distributions, the negative reference is essential here.
\end{example}
 
The contrast between the two examples previews the tradeoffs studied in this section: GEM demands the strongest model access (token-representation logits), whereas TVD-MI needs only a yes/no API call but relies on the reliability of the oracle's judgment.

The rest of this section formally instantiates the four design choices --- the information measure $f$-MI (\Cref{subsec:f-mi}), the MI estimation method (\Cref{subsec:mi-estimation}), the text representation (\Cref{subsec:representation}), and the prediction mechanism (\Cref{subsec:prediction}). \Cref{subsec:prior-work} then frames existing metrics as instances of the framework, and \Cref{subsec:manip-robust} shows that the resulting metrics inherit an approximate manipulation-robustness guarantee from information monotonicity, bounded by the estimation error.

\subsection{\texorpdfstring{$f$}{f}-Mutual Information}\label{subsec:f-mi}

In this paper, we instantiate the information-measure component \(F\) with the following two mutual-information objectives.
\begin{itemize}
    \item \textbf{KL mutual information}. Taking \(f_{\KL}(t)=t \log t\) gives the Shannon mutual information
    \[
        I_{\KL}(X;Y)
        =
        D_{\KL}(P_{XY}\|P_X\otimes P_Y)
        =
        \E_{\Pjoint}[\log r(X,Y)].
    \]
where $r(x, y) = \frac{p_{XY}(x,y)}{p_X(x)\,p_Y(y)}$ is the joint-to-product density ratio defined in \Cref{sec:preliminaries}.
    For a realized candidate--reference pair $(x_i,y_i)$, the sample-level score $\log r(x_i,y_i)$ is the pointwise mutual information used by GEM and GPPM-style metrics.

    \item \textbf{Total-variation mutual information}. Taking \(f_{\TV}(t)=\frac12|t-1|\) gives the total-variation mutual information (TVD-MI)
    \[
        I_{\TV}(X;Y)
        =
        \frac12\E_{\Pmarg}[|r(X,Y)-1|]
        =
        \frac12\sum_{x,y}|p_{XY}(x,y)-p_X(x)p_Y(y)|.
    \]
\end{itemize}

Our empirical design sweep therefore varies the information-measure component \(F\in\{\KL,\TV\}\) and crosses these two information measures with the estimation methods described next.

\subsection{MI Estimation}\label{subsec:mi-estimation}

Since the vast combinatorial space of text prevents the explicit expression of the joint candidate--reference distribution, the population objective $\fMI(X;Y)$ cannot be computed in closed form. To circumvent this, we explore two families of estimators.
\begin{itemize}
    \item \textbf{Direct Density Ratio Estimation}. 
By the density-ratio formulation of mutual information in \Cref{eq: mi density ratio}, if the density-ratio function $r$ is known, the $f$-mutual information can be calculated via a Monte Carlo estimator by averaging the transformed ratio $f(r(x,y))$ over candidate--reference samples $(x,y)$ drawn from the product-of-marginals distribution $P_X\otimes P_Y$. \Cref{apdx: estimation direct density ratio} provides examples of direct density-ratio estimation.
\item \textbf{$f$-Variational Estimation}. Another commonly used estimator in the literature is the $f$-variational estimator, constructed via the Fenchel conjugate $f^*$.
\begin{equation}
\label{eq: fvar}
    \fMI(X;Y)
    =
  \E_{\Pjoint}[T^*(X,Y)] - \E_{\Pmarg}\bigl[f^*(T^*(X,Y))\bigr] ,
\end{equation}
where the population-optimal critic $T^*(x,y) \in \partial f\bigl(r(x,y)\bigr)$ is a subgradient of the convex generator $f$.

The $f$-variational estimator takes a realized task $w_i$, candidate report $x_i$, positive reference $y_i$, and negative reference $y_i^-\sim P_Y$ drawn from the marginal reference distribution. The resulting sample-level score $S(x_i,y_i,y_i^-)$ is an unbiased estimator of the population objective. Specifically, the score replaces the expectation over the joint candidate--reference distribution $P_{XY}$ with the matched pair $(x_i,y_i)$ and the expectation over the product-of-marginals distribution $P_X\otimes P_Y$ with the mismatched pair $(x_i,y_i^-)$.
\end{itemize}

The two approaches differ in their estimation targets. Direct methods require estimating the density-ratio function $r$. Variational methods instead optimize a critic function $T$ under the dual objective and therefore do not need to construct an explicitly calibrated density-ratio estimate. The extent to which this target is simpler depends on the convex generator $f$. For example, for total-variation divergence, a population-optimal critic can be chosen as
\(
    T^*(x,y)
    =
    \frac{1}{2}\operatorname{sign}\bigl(r(x,y)-1\bigr),
\)
so only whether the density ratio lies above or below $1$ is required, rather than its magnitude. Thus, the requirement of the $f$-variational estimators is sometimes weaker. We defer further mathematical details to \Cref{apdx: estimation f variational}.

\paragraph{Examples.} Table~\ref{tab:fmi-what-estimate} summarizes the four cases and makes explicit the sample-level score $S(x_i,y_i,y_i^-)$ that the prediction mechanism $\Pi$ needs to estimate. Averaging these sample-level scores over the $n$ observed tasks estimates the corresponding population $f$-mutual information, as required by the design in Definition~\ref{def:triple}.

\begin{table}[htbp]
\centering
\small
\caption{Sample-level score and quantity to estimate for each \((F,E)\) pair.}
\label{tab:fmi-what-estimate}
\begin{tabular}{@{}llll@{}}
\toprule
\(F\) & \(E\) & Sample-level score & Quantity to estimate \\
\midrule
\(\KL\)
& Direct
& \(\log r(x_i,y_i)\)
& \(\log r(x_i,y_i)\) \\

\(\KL\)
& \(f\)-variational
& \(T^*(x_i,y_i)-e^{T^*(x_i,y_i^-)-1}\)
& \(T^*(x_i,y_i)=1+\log r(x_i,y_i)\) \\

\(\TV\)
& Direct
& \(\frac12\left|1-r(x_i,y_i)^{-1}\right|\) ~or~ \(\frac12\left|r(x_i,y_i^-)-1\right|\)
& \(r(x_i,y_i)\) ~or~ \(r(x_i,y_i^-)\)\\

\(\TV\)
& \(f\)-variational
& \(T^*(x_i,y_i)-T^*(x_i,y_i^-)\)
& \(T^*(x_i,y_i)=\frac12\,\operatorname{sign}(r(x_i,y_i)-1)\) \\
\bottomrule
\end{tabular}
\end{table}

Note that KL-direct and TV-direct\footnote{The two TV-direct forms have the same expectation:
\(
    \E_{\Pjoint}\left[
        \frac{1}{2}\left|1-r(X,Y)^{-1}\right|
    \right]
    =
    \E_{\Pmarg}\left[
        \frac{1}{2}\left|r(X,Y)-1\right|
    \right].
\)} both require estimating the density ratio $r(x,y)$, while the variational estimators require a critic $T^*(x,y)$. In particular, TV-variational only needs a bounded classifier for whether $r(x,y)$ is above or below $1$. 

Producing these estimators from text requires two further choices: a representation $R$ that fixes the granularity at which the report is processed (Section~\ref{subsec:representation}), and a prediction mechanism $\Pi$ that maps represented pairs to the required predictive quantities (Section~\ref{subsec:prediction}).

\subsection{Representation}\label{subsec:representation}

The representation $R:\mathcal{V}^{*}\to\mathcal{U}$ fixes the units at which a report is processed by the downstream predictor. We consider three natural granularities.

\begin{itemize}
    \item \textbf{Token}. At the finest granularity, a report $y$ is represented by its token sequence $(y_1,\ldots,y_T)$. This preserves all information in the text and exposes the autoregressive factorization of language models, $p(y\mid x)=\prod_{t=1}^{T}p(y_t\mid y_{<t},x)$, so that full-report-representation conditional probabilities can be assembled from per-token factors. Token representation is only meaningful in combination with a predictor that consumes token-representation signals, such as an autoregressive language model with logit access.

In practice, prior work~\citep{xu2025benchmarking,lu2024eliciting} reports that style-normalization pre-processing is important when using token representations, since autoregressive language models can be confounded by superficial features of the text such as writing style.

\item \textbf{Statement}. An intermediate granularity decomposes a report into its atomic claims. Let
\[
\Psi(y) = \bigl\{\psi_1(y),\ldots,\psi_{K(y)}(y)\bigr\}
\]
denote the statement decomposition of $y$, where each $\psi_k(y)\in\mathcal{V}^{*}$ is an atomic statement, typically obtained via a separate LLM call. Decomposing a report into statement-level representation naturally loses information by the data-processing inequality (\Cref{prop:dpi-fmi}). In practice, however, isolating individual claims often improves the predictor's performance, especially for an LLM-oracle, which tends to reason more reliably about single facts than about long passages. With statement-level decomposition, the estimation quality may even improve. We compare these tradeoffs empirically in Section~\ref{sec:factorial} and Section~\ref{sec:results}.

\item \textbf{Full-Report}. At the coarsest granularity, a report is treated as an atomic unit, $R(y)=y$. Predictive quantities are obtained holistically from the candidate-reference pair $(x,y)$: a single conditional likelihood $p(y\mid x)$, or a single oracle judgment about the pair. (e.g., ``do these two reports describe the same item?'').
\end{itemize}

\subsection{Prediction Mechanism}\label{subsec:prediction}

The prediction mechanism $\Pi$ maps a represented candidate-reference pair to the predictive quantities in the representation space, as required by the chosen estimator $E$. See \Cref{tab:fmi-what-estimate} for examples of required predictive quantities. Below lists potential prediction mechanisms. 

\begin{itemize}
    \item \textbf{Autoregression}. An autoregressive language model with parameters $\phi$ assigns a probability to any token sequence via the chain rule. For a candidate report $x$ in the prompt, we obtain
\[
p_\phi(y \mid x) = \prod_{t=1}^{T} p_\phi(y_t \mid y_{<t}, x)
\]
where each factor is read off the softmax over output logits. 
Autoregression interacts with the \emph{token} level representation. It also requires logit access, which some commercial APIs do not expose, and inherits any miscalibration of the underlying language model. As an example, for KL-based MI estimation, substituting a null report for $x$ yields a marginal estimate $p_\phi(y)$, and the following PMI is obtained for direct KL estimation:
\[
\widehat{\PMI}(x;y) = \log r(x,y) = \log p_\phi(y\mid x) - \log p_\phi(y).
\]
Exponentiating gives a density-ratio estimate that feeds any of the four $(F,E)$ pairs in Table~\ref{tab:fmi-what-estimate}.
\item \textbf{LLM-Oracle}.
When token-representation probabilities are unavailable, an LLM can be prompted for a prediction about a represented pair. The prediction is then used as inputs to the estimators as required. The following two regimes of an LLM oracle cover the cases in Table~\ref{tab:fmi-what-estimate}:
\begin{itemize}
    \item \emph{Critic-style prediction} fits naturally into the $f$-variational framework: a scalar or categorical LLM is prompted to directly output a critic used as $T^*(x, y)$. For example, \citet{robertson2025beyondvibes} shows that the TVD-MI is especially convenient. The $T^*$ required in \Cref{eq: fvar} for TVD-MI can be implemented as a binary classifier distinguishing same-task from cross-task pairs. A simple prompt, e.g., ``Do these two reports describe the same item?'' implements this $T^*$ for TVD-MI.
    \item \emph{Likelihood-style prediction} supports direct estimation. The LLM is asked, for instance, how strongly the candidate $x$ supports the reference $y$, using a small set of ordered probability or support bins~\citep{lu2024eliciting}. Mapping the selected bin to a numeric value yields a coarse surrogate for $\log \hat p(y\mid x) - \log \hat p(y)$, and hence for the density ratio $\hat r(x,y)$. Such direct LLM-oracle estimators are typically biased. Their outputs are discretized and often poorly calibrated. But they provide a workable approximation when only API access is available.
\end{itemize}

LLM-oracle prediction composes with the full-report and statement representations. At the \emph{full-report} level, the prompt simply contains $x$ and $y$ and the oracle returns a single prediction. At the \emph{statement} level, the candidate report $x$ is paired with each $\psi_k(y)$, and the resulting per-statement quantities are combined into a full-report-representation prediction by an aggregation rule
\(
A:\bigcup_{K\ge 1}\mathbb{R}^K\to\mathbb{R}.
\)

\end{itemize}

\subsection{Prior Work}
\label{subsec:prior-work}

Combining the preceding components, a complete design tuple $(F,E,R,\Pi)$ specifies a concrete reference-based metric. Table~\ref{tab:design-coverage} shows how three existing methods instantiate this framework.

\begin{table}[htbp]
\centering
\caption{Coverage of the design space. Each cell is labeled with the name of the corresponding evaluation metric, when one exists. \textsc{new}  denotes combinations introduced or evaluated in this work;
}
\footnotesize
\label{tab:design-coverage}
\renewcommand{\arraystretch}{1.25}
\begin{tabular}{@{}llccc@{}}
\toprule
& & \multicolumn{1}{c}{$\Pi$ = \textbf{Autoregression}}
  & \multicolumn{2}{c}{$\Pi$ = \textbf{LLM-oracle}} \\
\cmidrule(lr){3-3}\cmidrule(lr){4-5}
$F$ & $E$ & $R$ = Token & $R$ = Full Report & $R$ = Statement \\
\midrule
\multirow{2}{*}{KL}
    & Direct          & GEM \citep{xu2025benchmarking}       & \textsc{new} & GPPM-J \citep{lu2024eliciting} \\
    & $f$-variational & \textsc{new} & \textsc{new}   & \textsc{new}           \\
\midrule
\multirow{2}{*}{TV}
    & Direct          & \textsc{new}    & \textsc{new} & \textsc{new}            \\
    & $f$-variational & \textsc{new}    & TVD-MI \citep{robertson2025beyondvibes} & \textsc{new} \\
\bottomrule
\end{tabular}
\end{table}

A practical benefit of this decomposition is that it makes unexplored combinations easy to identify.
Cells of the tuple $(F,E,R,\Pi)$ that Table~\ref{tab:design-coverage} 
with \textsc{new} correspond to metrics that are implementable in principle but have not been studied. In Section~\ref{sec:factorial}, we empirically study several such metrics, as shown in Table~\ref{tab:design-coverage}.

The practical benefit of the framework is that it exposes new metrics. For example, TVD-MI \citep{robertson2025beyondvibes} uses the tuple
\(
(\TV,\ f\text{-variational},\ \text{Full-report},\ \text{LLM-oracle}).
\)
Changing the representation gives the statement-level TV
\(f\)-variational metric (formally defined in Appendix~\ref{sec:contrastive-tv}):
\[
(\TV,\ f\text{-variational},\ \text{Statement},\ \text{LLM-oracle}).
\]
This metric decomposes the positive and negative references into atomic statements, asks the LLM critic whether each statement is supported by the candidate, and compares support for the true reference statement against support for the negative reference statement. As a result, candidate content that supports statements in the matched reference is rewarded, while support for statements drawn from an unrelated task is penalized. Generic content is discounted because it tends to support both the positive and negative references. Thanks to the finer-grained representation, in our experiment in Section~\ref{sec:factorial}, this metric achieves better strategic and statistical alignment than TVD-MI across multiple datasets.

\subsection{Manipulation Robustness of the \texorpdfstring{$f$}{f}-MI Metrics}
\label{subsec:manip-robust}

The data-processing inequality in Proposition~\ref{prop:dpi-fmi} guarantees that any candidate manipulation
\(\strategy:\mathcal{X}\to\mathcal{X}\) can only reduce the
\emph{true} $f$-MI:
\[
\fMI(\strategy(X);Y)\le \fMI(X;Y).
\]
However, the score is computed using a fixed estimator and distribution model, which in practice amounts to using a restricted or suboptimal critic rather than the population optimum. As a result, the estimated score need not satisfy DPI exactly, even when the underlying information measure does. This motivates the following approximate notion of robustness: for a manipulation class \(\Sigma\), suppose
\begin{equation*}
\sup_{\strategy\in\Sigma}
\left|
\E\left[S(\strategy(X),Y)\right]
-
\fMI(\strategy(X);Y)
\right|
\le \varepsilon.
\end{equation*}
Then for every candidate manipulation \(\strategy\in\Sigma\),
\begin{equation*}
\E\left[S(\strategy(X),Y)\right]
\le
\E\left[S(X,Y)\right] + 2\varepsilon.
\end{equation*}
Indeed,
\[
\E\left[S(\strategy(X),Y)\right]
\le
\fMI(\strategy(X);Y)+\varepsilon
\le
\fMI(X;Y)+\varepsilon
\le
\E\left[S(X,Y)\right]+2\varepsilon,
\]
where the middle inequality is exactly Proposition~\ref{prop:dpi-fmi}. Thus, any manipulation that decreases the true $f$-mutual information by more than twice the uniform estimation-error bound, \(2\varepsilon\), is still penalized by the estimated score in expectation.

The assumption requires the estimation error to be uniformly small over the entire manipulation class $\Sigma$. This is a strong condition: it asks the critic to generalize not only to clean reports but also to all manipulated variants. In practice, the uniform error bound $\varepsilon$ may be large for manipulation strategies that produce out-of-distribution inputs for the critic. Empirical validation (Section~\ref{sec:factorial}) is therefore essential.

\section{Experiment}
\label{sec:factorial}

We compare evaluation metrics in a controlled way under our test principles (\Cref{sec:eval-pipeline}), with detailed setup deferred to \Cref{apdx: experiment design}. The code is provided in \url{https://github.com/DavidXu999/Statistical-and-Strategic-Alignment-for-Text-Evaluation-Metrics}.

\paragraph{Research Question.} 
To keep the analysis focused, we frame the experiments around the decision a practitioner actually faces. 
The predictor $\Pi$ and representation $R$ are typically dictated by task and deployment constraints, e.g., whether logit access is available, and whether atomic statements can be reliably extracted, while the information measure $F$ and estimator $E$ are under the researcher's direct control.
We therefore ask:

\emph{Given a fixed predictor and representation pair $(\Pi, R)$, which information measure and estimator $(F, E)$ yields the best-behaved evaluation metric?}

\paragraph{Other reference-based baselines.}
We also compare against popular reference-based metrics as baselines, including ROUGE-L, BLEU, BERTScore (based on roberta-large), and LLM-as-a-Judge (based on various models). These baselines test whether accuracy and manipulation-resistance gains from the MI framework go beyond lexical overlap, embedding similarity, or direct LLM judging.

\paragraph{Datasets.}
We evaluate metrics across three domains that stress different aspects of reference-based text evaluation: peer review, summarization, and question answering. In peer review, the task input is a paper or essay and the candidate reports are full reviews; in summarization, the task input is a source document and the reports are summaries; in question answering, the task input is a question and the reports are answers. Across these domains, reports differ substantially in length, structure, subjectivity, and semantic complexity. 

Our evaluation includes seven datasets: three peer-review datasets, two summarization datasets, and two question-answering datasets. Four datasets provide item-level human quality annotations and are used to measure human-rating correlation; all datasets are used for degradation and manipulation tests. When multiple independent reports are available for the same task, we treat them as positive references. 
Table~\ref{tab:datasets} summarizes the datasets.\footnote{Peer grading-WH \citep{wu2024elicitationgpt} and Peer grading-XLSK \citep{xu2025benchmarking} are not publicly accessible while the others are public datasets.}

\begin{table}[htbp]
\centering
\caption{Overview of evaluation datasets. $n$ is the number of candidate reports; \#ref is the typical number of positive references per task. ``Human rating'' indicates whether item-level human quality annotations are available for candidate reports.}
\label{tab:datasets}
\renewcommand{\arraystretch}{1.18}
\begin{tabular}{@{}lllccc@{}}
\toprule
\textbf{Domain} & \textbf{Dataset} & $n$ & Ref & \textbf{Human rating} & \textbf{Reference} \\
\midrule
\multirow{3}{*}{Peer review}
    & ICLR 2026            & $300^*$ & peer & No & OpenReview \\
    & Peer grading-WH       & $534$  & GT & Numerical & \citet{wu2024elicitationgpt} \\
    & Peer grading-XLSK     & $165$  & peer & Numerical & \citet{xu2025benchmarking} \\
\midrule
\multirow{2}{*}{Summarization}
    & SummEval             & $800^*$  &  GT    & Numerical & \citet{fabbri2021summeval} \\
    & SPACE               & $50$ &  GT   & No & \citet{angelidis2021extractive} \\
\midrule
\multirow{2}{*}{QA}
    & LFQA-E               & $320^*$ &  GT   & Pairwise & \citet{fan2026lfqa} \\
    & MedAESQA             & $400^*$ &  GT   & Numerical & \citet{gupta2025dataset} \\
\bottomrule
\end{tabular}
 
\vspace{2pt}
{\footnotesize $^*$~Down-sampled from the whole dataset for a more tractable test scale.}
\end{table}

\subsection{Results}\label{sec:results}

\begin{table}[htbp]
\centering
\caption{Human-rating correlation results. Rows are evaluation metrics, grouped by distribution model $D$ and its granularity; within each group, rows sweep the information measure $F\in\{\KL, \TV\}$ and estimator $E\in\{\text{Direct}, \text{$f$-var.}\}$. Columns are datasets; entries are Spearman rank correlations with human ratings. For SummEval we report the mean over all dimensions. Best result within each group is \textbf{bolded}; best among $f$-MI metrics per column and best overall per column are \underline{underlined}.}
\label{tab:correlation-main}
\renewcommand{\arraystretch}{0.85}
\setlength{\tabcolsep}{7pt}

\begin{tabular}{@{}lcccc@{}}
\toprule
\textbf{Evaluation Metric}
& \multicolumn{2}{c}{\textbf{Peer review}}
& \multicolumn{1}{c}{\textbf{Summarization}}
& \multicolumn{1}{c}{\textbf{QA}} \\
\cmidrule(lr){2-3}\cmidrule(lr){4-4}\cmidrule(lr){5-5}
$(F, E)$
    & PG-WH & PG-XLSK
    & SummEval
    & MedAESQA \\
\midrule
\multicolumn{5}{@{}l}{\textit{Autoregression, Token-representation}} \\
\midrule
$(\KL, \text{Direct})$ (GEM) & \underline{\textbf{0.460}} & \underline{\textbf{0.475}} & \textbf{0.138} & 0.113 \\
$(\KL, \text{$f$-var.})$ & -0.218 & -0.355 & 0.059 & 0.051 \\
$(\TV, \text{Direct})$ & 0.312 & 0.380 & 0.095 & -0.025 \\
$(\TV, \text{$f$-var.})$ & -0.053 & -0.036 & 0.085 & \textbf{0.129} \\
\midrule
\multicolumn{5}{@{}l}{\textit{LLM-oracle, Report-representation}} \\
\midrule
$(\KL, \text{Direct})$ & 0.235 & 0.367 & 0.031 & \textbf{0.163} \\
$(\KL, \text{$f$-var.})$ & 0.252 & \textbf{0.419} & 0.031 & 0.162 \\
$(\TV, \text{Direct})$ & -0.231 & -0.350 & 0.017 & -0.238 \\
$(\TV, \text{$f$-var.})$ (TVD-MI) & \textbf{0.286} & 0.042 & \textbf{0.140} & 0.110 \\
\midrule
\multicolumn{5}{@{}l}{\textit{LLM-oracle, Statement-representation}} \\
\midrule
$(\KL, \text{Direct})$ (GPPM-J) & 0.254 & 0.307 & 0.073 & 0.041 \\
$(\KL, \text{$f$-var.})$ & \textbf{0.281} & 0.251 & 0.075 & 0.064 \\
$(\TV, \text{Direct})$ & -0.292 & -0.396 & -0.043 & -0.129 \\
$(\TV, \text{$f$-var.})$ & 0.201 & \textbf{0.468} & \underline{\textbf{0.208}} & \underline{\textbf{0.228}} \\
\midrule
\midrule
\multicolumn{5}{@{}l}{\textit{Non-MI baselines}} \\
\midrule
ROUGE-L & -0.244 & 0.167 & 0.111 & 0.151 \\
BLEU & -0.332 & 0.256 & 0.102 & 0.127 \\
BERTScore & -0.133 & -0.063 & 0.252 & 0.166 \\
LLM-as-Judge / Claude-Haiku-4.5 & 0.462 & 0.569 & 0.400 & \underline{\textbf{0.315}} \\
LLM-as-Judge / Claude-Sonnet-4.5 & 0.512 & 0.622 & 0.450 & 0.301 \\
LLM-as-Judge / GPT-5-mini & \underline{\textbf{0.539}} & \underline{\textbf{0.631}} & \underline{\textbf{0.477}} & 0.256 \\
LLM-as-Judge / GPT-4o-mini & 0.492 & 0.528 & 0.452 & 0.208 \\
\bottomrule
\end{tabular}
 \end{table}

Table~\ref{tab:correlation-main} reports Spearman correlation with human ratings on the four datasets carrying item-level numerical annotations: Peer Grading-WH \citep{wu2024elicitationgpt}, Peer Grading-XLSK \citep{xu2025benchmarking}, SummEval \citep{fabbri2021summeval}, and MedAESQA \citep{gupta2025dataset}. Rows are grouped by the representation and predictor pair $(R,\Pi)$; within each block, the four rows sweep $(F,E)\in\{\KL,\TV\}\times\{\text{Direct},\text{$f$-var.}\}$. Bold marks the best method within each block; underlines mark the best $f$-MI metric and the best overall metric per dataset.

\begin{table}[htbp]
\centering
\scriptsize
\setlength{\tabcolsep}{3pt}
\renewcommand{\arraystretch}{1.1}
\caption{Sensitivity to degradation. For each (metric, dataset) pair, we apply $m$ degradation strategies (header) and count those on which the metric \emph{fails} to produce a significant score decrease at $p<0.05$; lower is better. A \checkmark{} marks zero failures; otherwise, letter codes mark the strategies the metric missed (legend below).}
\label{tab:degradation-summary}
\resizebox{\linewidth}{!}{\begin{tabular}{@{}lrrrrrrrr@{}}
\toprule
\textbf{Evaluation Metric}& \multicolumn{3}{c}{\textbf{Peer review}}& \multicolumn{2}{c}{\textbf{Summarization}}& \multicolumn{2}{c}{\textbf{QA}}& \textbf{total} \\
\cmidrule(lr){2-4}\cmidrule(lr){5-6}\cmidrule(lr){7-8}
$(F, E)$
    & PG-WH & PG-XLSK & ICLR26 & SummEval & SPACE & LFQA-E & MedAESQA & fail \\
    & $m=5$ & $m=6$ & $m=6$ & $m=5$ & $m=5$ & $m=5$ & $m=5$ & $m=37$ \\
\midrule
\multicolumn{9}{@{}l}{\textit{MI-based Mechanisms}} \\
KL-Direct-Autoreg. & $\checkmark$ & $\checkmark$ & $\textcolor{orange}{\mathrm{P}}$ & $\checkmark$ & $\checkmark$ & $\textcolor{orange}{\mathrm{P}}\,\textcolor{orange}{\mathrm{U}}$ & $\textcolor{orange}{\mathrm{P}}\,\textcolor{orange}{\mathrm{U}}$ & 5 \\
TVD-FVar-Report & $\textcolor{orange}{\mathrm{D}}\,\textcolor{orange}{\mathrm{S}}\,\textcolor{orange}{\mathrm{U}}$ & $\textcolor{orange}{\mathrm{O}}$ & $\textcolor{orange}{\mathrm{O}}\,\textcolor{orange}{\mathrm{P}}\,\textcolor{orange}{\mathrm{U}}$ & $\textcolor{orange}{\mathrm{P}}\,\textcolor{red}{\mathrm{U}}$ & $\checkmark$ & $\textcolor{orange}{\mathrm{P}}\,\textcolor{orange}{\mathrm{U}}$ & $\textcolor{red}{\mathrm{D}}\,\textcolor{orange}{\mathrm{S}}\,\textcolor{orange}{\mathrm{P}}\,\textcolor{red}{\mathrm{U}}$ & 15 \\
TVD-FVar-Statement & $\checkmark$ & $\checkmark$ & $\checkmark$ & $\textcolor{orange}{\mathrm{P}}$ & $\checkmark$ & $\textcolor{orange}{\mathrm{P}}$ & $\textcolor{orange}{\mathrm{P}}$ & 3 \\
\midrule
\multicolumn{9}{@{}l}{\textit{Non-MI baselines}} \\
LLM-as-Judge / Claude-Haiku-4.5 & $\checkmark$ & $\checkmark$ & $\textcolor{red}{\mathrm{P}}$ & $\textcolor{red}{\mathrm{P}}$ & $\checkmark$ & $\textcolor{red}{\mathrm{P}}$ & $\textcolor{orange}{\mathrm{D}}\,\textcolor{red}{\mathrm{P}}$ & 5 \\
LLM-as-Judge / Claude-Sonnet-4.5 & $\checkmark$ & $\checkmark$ & $\textcolor{orange}{\mathrm{P}}$ & $\textcolor{red}{\mathrm{P}}$ & $\checkmark$ & $\textcolor{red}{\mathrm{P}}$ & $\textcolor{red}{\mathrm{P}}$ & 4 \\
LLM-as-Judge / GPT-4o-mini & $\textcolor{orange}{\mathrm{D}}$ & $\textcolor{red}{\mathrm{D}}\,\textcolor{red}{\mathrm{O}}\,\textcolor{orange}{\mathrm{R}}$ & $\textcolor{orange}{\mathrm{O}}\,\textcolor{orange}{\mathrm{R}}\,\textcolor{red}{\mathrm{P}}$ & $\textcolor{red}{\mathrm{P}}$ & $\checkmark$ & $\textcolor{red}{\mathrm{P}}$ & $\textcolor{red}{\mathrm{P}}$ & 10 \\
LLM-as-Judge / GPT-5-mini & $\checkmark$ & $\checkmark$ & $\textcolor{orange}{\mathrm{P}}$ & $\textcolor{orange}{\mathrm{P}}$ & $\checkmark$ & $\textcolor{red}{\mathrm{P}}$ & $\textcolor{orange}{\mathrm{D}}\,\textcolor{red}{\mathrm{P}}$ & 5 \\
\addlinespace[0.35em]
\multicolumn{9}{@{}l}{\footnotesize Codes: D=deletion and completion; O=opinion flip; R=random replacement.} \\
\multicolumn{9}{@{}l}{\footnotesize Codes: S=sentence deletion; P=surface report; U=ultra concise compression.} \\
\multicolumn{9}{@{}l}{\footnotesize \textcolor{red}{Red}: significant score increase. \textcolor{orange}{Orange}: non-significant change.} \\
\bottomrule
\end{tabular}
 }
\end{table}
\begin{table}[htbp]
\centering
\scriptsize
\setlength{\tabcolsep}{3pt}
\renewcommand{\arraystretch}{1.1}
\caption{Manipulation resistance. For each (metric, dataset) pair, we apply $m$ manipulation strategies (header) and count those the metric \emph{fails} to resist, i.e., that produce a significant score increase at $p<0.05$; lower is better. A \checkmark{} marks zero failures; otherwise, letter codes flag the strategies the metric failed to resist (legend below).}
\label{tab:manipulation-summary}
\resizebox{\linewidth}{!}{\begin{tabular}{@{}lrrrrrrrr@{}}
\toprule
\textbf{Evaluation Metric}& \multicolumn{3}{c}{\textbf{Peer review}}& \multicolumn{2}{c}{\textbf{Summarization}}& \multicolumn{2}{c}{\textbf{QA}}& \textbf{total} \\
\cmidrule(lr){2-4}\cmidrule(lr){5-6}\cmidrule(lr){7-8}
$(F, R, \Pi, E)$
    & PG-WH & PG-XLSK & ICLR26 & SummEval & SPACE & LFQA-E & MedAESQA & fail \\
    & $m=6$ & $m=6$ & $m=6$ & $m=2$ & $m=2$ & $m=4$ & $m=4$ & $m=30$ \\
\midrule
\multicolumn{9}{@{}l}{\textit{MI-based Mechanisms}} \\
(KL, Token, Autoregression, Direct) & $\textcolor{red}{\mathrm{M}}$ & $\textcolor{red}{\mathrm{S}}$ & $\textcolor{red}{\mathrm{N}}\,\textcolor{red}{\mathrm{H}}\,\textcolor{red}{\mathrm{P}}\,\textcolor{red}{\mathrm{S}}\,\textcolor{red}{\mathrm{R}}$ & $\checkmark$ & $\checkmark$ & $\checkmark$ & $\checkmark$ & 7 \\
(TV, Report, LLM-oracle, $f$-var) & $\checkmark$ & $\checkmark$ & $\checkmark$ & $\checkmark$ & $\checkmark$ & $\checkmark$ & $\checkmark$ & \textbf{0} \\
(TV, Statement, LLM-oracle, $f$-var) & $\checkmark$ & $\checkmark$ & $\checkmark$ & $\checkmark$ & $\checkmark$ & $\checkmark$ & $\checkmark$ & \textbf{0} \\
\midrule
\multicolumn{9}{@{}l}{\textit{Non-MI baselines}} \\

LLM-as-Judge / Claude-Haiku-4.5 & $\textcolor{red}{\mathrm{M}}\,\textcolor{red}{\mathrm{N}}\,\textcolor{red}{\mathrm{H}}\,\textcolor{red}{\mathrm{S}}\,\textcolor{red}{\mathrm{R}}$ & $\textcolor{red}{\mathrm{H}}\,\textcolor{red}{\mathrm{P}}\,\textcolor{red}{\mathrm{S}}\,\textcolor{red}{\mathrm{R}}$ & $\textcolor{red}{\mathrm{H}}\,\textcolor{red}{\mathrm{S}}\,\textcolor{red}{\mathrm{R}}$ & $\textcolor{red}{\mathrm{R}}$ & $\textcolor{red}{\mathrm{R}}$ & $\textcolor{red}{\mathrm{H}}\,\textcolor{red}{\mathrm{S}}\,\textcolor{red}{\mathrm{R}}$ & $\textcolor{red}{\mathrm{H}}\,\textcolor{red}{\mathrm{S}}\,\textcolor{red}{\mathrm{R}}$ & 20 \\
LLM-as-Judge / Claude-Sonnet-4.5 & $\textcolor{red}{\mathrm{M}}\,\textcolor{red}{\mathrm{N}}\,\textcolor{red}{\mathrm{H}}\,\textcolor{red}{\mathrm{S}}\,\textcolor{red}{\mathrm{R}}$ & $\textcolor{red}{\mathrm{H}}\,\textcolor{red}{\mathrm{P}}\,\textcolor{red}{\mathrm{S}}\,\textcolor{red}{\mathrm{R}}$ & $\textcolor{red}{\mathrm{M}}\,\textcolor{red}{\mathrm{H}}\,\textcolor{red}{\mathrm{P}}\,\textcolor{red}{\mathrm{S}}\,\textcolor{red}{\mathrm{R}}$ & $\textcolor{red}{\mathrm{R}}$ & $\checkmark$ & $\textcolor{red}{\mathrm{H}}\,\textcolor{red}{\mathrm{S}}$ & $\textcolor{red}{\mathrm{H}}$ & 18 \\
LLM-as-Judge / GPT-4o-mini & $\textcolor{red}{\mathrm{M}}\,\textcolor{red}{\mathrm{N}}\,\textcolor{red}{\mathrm{H}}\,\textcolor{red}{\mathrm{P}}\,\textcolor{red}{\mathrm{S}}\,\textcolor{red}{\mathrm{R}}$ & $\textcolor{red}{\mathrm{M}}\,\textcolor{red}{\mathrm{N}}\,\textcolor{red}{\mathrm{H}}\,\textcolor{red}{\mathrm{P}}\,\textcolor{red}{\mathrm{S}}\,\textcolor{red}{\mathrm{R}}$ & $\textcolor{red}{\mathrm{M}}\,\textcolor{red}{\mathrm{N}}\,\textcolor{red}{\mathrm{H}}\,\textcolor{red}{\mathrm{P}}\,\textcolor{red}{\mathrm{S}}\,\textcolor{red}{\mathrm{R}}$ & $\textcolor{red}{\mathrm{R}}$ & $\textcolor{red}{\mathrm{R}}$ & $\textcolor{red}{\mathrm{H}}\,\textcolor{red}{\mathrm{S}}\,\textcolor{red}{\mathrm{R}}$ & $\textcolor{red}{\mathrm{H}}\,\textcolor{red}{\mathrm{S}}\,\textcolor{red}{\mathrm{R}}$ & 26 \\
LLM-as-Judge / GPT-5-mini & $\textcolor{red}{\mathrm{N}}\,\textcolor{red}{\mathrm{H}}\,\textcolor{red}{\mathrm{S}}\,\textcolor{red}{\mathrm{R}}$ & $\textcolor{red}{\mathrm{M}}\,\textcolor{red}{\mathrm{N}}\,\textcolor{red}{\mathrm{H}}\,\textcolor{red}{\mathrm{P}}\,\textcolor{red}{\mathrm{S}}\,\textcolor{red}{\mathrm{R}}$ & $\textcolor{red}{\mathrm{H}}\,\textcolor{red}{\mathrm{P}}\,\textcolor{red}{\mathrm{S}}\,\textcolor{red}{\mathrm{R}}$ & $\checkmark$ & $\checkmark$ & $\textcolor{red}{\mathrm{H}}\,\textcolor{red}{\mathrm{S}}\,\textcolor{red}{\mathrm{R}}$ & $\textcolor{red}{\mathrm{H}}\,\textcolor{red}{\mathrm{S}}$ & 19 \\
\addlinespace[0.35em]
\multicolumn{9}{@{}l}{\footnotesize Codes: M=meaningless elongation; N=negative opinion shift; H=hedged opinion shift.} \\
\multicolumn{9}{@{}l}{\footnotesize Codes: P=positive opinion shift; S=strong opinion shift; R=rephrase.} \\
\multicolumn{9}{@{}l}{\footnotesize \textcolor{red}{Red} codes mark significant score increases after manipulation.} \\
\bottomrule
\end{tabular}
 }
\end{table}

For degradation and manipulation tests, a full sweep of all configurations across every perturbation strategy and dataset is computationally prohibitive. We therefore carry forward the best $f$-MI configurations from each representation-prediction block in Table~\ref{tab:correlation-main}: $(\KL,\text{Direct})$ with token-representation autoregression, corresponding to GEM; and $(\TV,\text{$f$-var.})$ with an LLM-oracle at both report and statement levels. This selection deliberately favors the correlation winners and asks whether they remain reliable under strategic perturbations. For large datasets, we down-sample to roughly 200 candidate reports per setting. We report result overviews in Tables~\ref{tab:degradation-summary} and~\ref{tab:manipulation-summary}, and detailed statistics in \Cref{app:degradation-manipulation-statistics}.

\paragraph{Key takeaways.}
The results show a separation between statistical and strategic alignment.

\begin{description}
    \item[\textbf{LLM-as-a-Judge is strongest on human-rating correlation, but collapses under manipulation.}]
    Among the non-MI baselines, LLM-as-a-Judge achieves the strongest correlations with human ratings. By the standard statistical alignment alone, it appears to be the best metric.  However, across the four judge models, manipulations succeed in 18 to 26 of the 30 tests. 
    \item[\textbf{The best $f$-MI design depends on the representation and predictor.}]
    Under token-representation autoregression, $(\KL,\text{Direct})$ (GEM) leads its block on three of four datasets and is the best $f$-MI metric overall on both peer-review datasets. Under an LLM-oracle, statement-level representation is substantially more effective: statement-level $(\TV,\text{$f$-var.})$ outperforms full-report-representation TVD-MI on three of four datasets and is the best $f$-MI metric overall on SummEval and MedAESQA.
    \item[\textbf{The unified design framework identifies a dominating new metric.}]
    The statement-level $(\TV,\text{$f$-var.})$ metric is not present in prior work, but is exposed by our design framework. Its strong performance suggests that decomposing references into atomic claims can make the oracle prediction task easier, enough to compensate for the information loss introduced by statement extraction. The $(\TV,\text{$f$-var.})$ metric is the most reliable on degradation sensitivity and manipulation robustness. It fails only 3 of 37 degradation tests, compared with 5 failures for GEM and 15 failures for full-report-representation $(\TV,\text{$f$-var.})$ / TVD-MI. It has 0 failures out of 30 manipulation tests, meaning that no tested manipulation produces a significant score increase.
    \item[\textbf{Robustness of MI-based metrics comes from the design, not merely from the model.}]
    Since Claude-Haiku-4.5 is vulnerable as an absolute LLM judge but robust when used inside the TV $f$-variational metric, the improvement comes from the MI-based formulation. 
\end{description}

It is worth noticing that no single axis of $(F,E,R,\Pi)$ dominates in isolation; rather, the design choices interact. token-representation autoregression favors KL-direct estimation, while LLM-oracle prediction benefits from the TV $f$-variational formulation, especially with statement-level representations. MI-based metrics also appear stronger on peer-review tasks than on summarization and QA, where the evaluation target can be more complex or subjective.

ROUGE-L, BLEU, and BERTScore are generally weaker baselines, and each fails to obtain positive correlation on at least one dataset, especially in peer review. This suggests that lexical overlap and embedding similarity are insufficient for evaluating richer, more subjective reports.

\paragraph{Additional discussion.} 
In ICLR2026, GEM fails the rephrase manipulation test, which appears to contrast with the original GEM results \citep{xu2025benchmarking}, where rephrasing did not produce a significant score increase in ICLR2023. A likely hypothesis is that our rephrase manipulation is stronger: Whereas \citet{xu2025benchmarking} primarily prompt the model to improve language quality, our prompt also allows the model to reorganize the review. Although it prohibits adding new facts or changing the reviewer's stance, this reorganization can make the candidate review easier for the metric to match with the reference. The result therefore suggests that GEM retains some sensitivity to presentation quality. 

However, the standardized mean difference for GEM remains relatively small ($+0.07 \pm 0.05$, 95\% CI), especially compared with direct LLM judging, even for the two most robust judge models: Claude-Sonnet-4.5 ($+0.25 \pm 0.09$, 95\% CI) and GPT-5-mini ($+0.17 \pm 0.10$, 95\% CI), as shown in Appendix~\ref{app:degradation-manipulation-statistics}. This observation reinforces the value of the MI-based formulation, and further motivates the new statement-level (TV, $f$-var.) metric, which remains robust to this stronger rephrasing test.

We also noticed that surface-report degradation is a strong test because the degraded reports are still fluent and task-specific, generated from weak but informative context such as a news lead, question, or paper abstract, by Claude-Haiku-4.5. Thus, they can remain plausible even while sometimes omitting important task-relevant information.

Under this test, MI-based metrics also compare favorably with the most robust LLM judges. MI-based metrics routinely pass more tests than LLM judges. Moreover, as shown in Appendix~\ref{app:degradation-manipulation-statistics}, all MI-based metrics either significantly decrease on surface reports or remain close to insensitive, with no significant positive failures. By contrast, even the strongest LLM judges do not penalize but reward surface reports: Claude-Sonnet-4.5 gives $d=+0.28\pm0.15$ (95\% CI) on SummEval, $+0.69\pm0.15$ on LFQA-E, and $+1.18\pm0.26$ on MedAESQA, while GPT-5-mini gives $+0.12\pm0.17$, $+0.51\pm0.14$, and $+1.01\pm0.19$ on the same datasets. Thus, surface reports expose a hard case for all metrics, but the failure mode differs: MI-based metrics can sometimes be under-sensitive, whereas LLM judges can directionally fail by rewarding plausible surface-level reports.

\section{Related Work}
\label{sec: related work}

\paragraph{NLG Evaluation}
Evaluation of AI systems spans many dimensions. Holistic frameworks consider accuracy, robustness, fairness, and toxicity~\citep{liang2022holistic}; alignment research emphasizes helpfulness, honesty, and harmlessness~\citep{askell2021general}; and a growing body of work targets truthfulness~\citep{lin2022truthfulqa}, the multi-faceted nature of fairness~\citep{gallegos2024bias}, and pluralism over legitimate human perspectives~\citep{sorensen2024roadmap}. 

Our framework instead targets a complementary and foundational property, \emph{semantic informativeness}: whether an evaluation score faithfully reflects how much useful, truthful, task-relevant information a report conveys. A response cannot be genuinely helpful if it carries little task-relevant signal, and mechanisms that reward truthful, informative reporting can discourage guessing and some forms of hallucination~\citep{wu2024elicitationgpt}.

\paragraph{Mutual Information Estimation}

Machine learning has produced sample-based MI estimators built on the variational representation of $f$-divergences via Fenchel duality \citep{nguyen2010estimating}, and neural network estimators such as MINE \citep{belghazi2018mine}, Deep InfoMax \citep{hjelm2018learning}, and InfoNCE \citep{oord2018representation}. 

In information elicitation without verification, MI plays a different role: \citet{kong2019information} score each agent a measure of MI between her report and a peer's reference, using the data-processing inequality to make truthful reporting an equilibrium. This echoes the same strategic-alignment principle behind our MI-based metrics (Proposition~\ref{prop:dpi-fmi}). \citet{kong2018water} extends the same MI-maximization principle to co-training, where two Bayesian predictors over distinct views of the data play the role of the two agents.

Closest to our setting are pre-trained language model approaches to MI estimation: \citet{padmakumar2021unsupervised} compute pointwise MI between extractive summaries and source articles, while \citet{xu2025benchmarking,lu2024eliciting,robertson2025beyondvibes} develop reference-based MI metrics for NLG evaluation, which can be recovered as specific instances by our design framework. Recent work also uses estimated MI for dataset evaluation \citep{chen2025data,zheng2024proper}.

\paragraph{Information Elicitation.}
Our notion of strategic alignment is motivated by the information elicitation literature, which designs mechanisms that induce agents to truthfully reveal private information. Two central lines are proper scoring rules and peer prediction. Proper scoring rules elicit truthful probabilistic predictions when a ground-truth outcome is eventually observed \citep{mccarthy1956measures, savage1971elicitation, gneiting2007strictly, lambert2008eliciting}. They are distinct from the evaluation metrics studied here, which score textual reports against references; the connection is that both shape the behavior of systems that optimize their scores. Peer prediction instead handles settings without direct verification, using the statistical relationship among peer reports as a substitute for ground truth \citep{miller2005eliciting, dasgupta2013crowdsourced, kong2019information,kong2020dominantly,kong2024dominantly,zhang2026stochastically}. Our setting parallels peer prediction because both use statistical relationships among reports when direct verification is unavailable. As \citet{burrell2021measurement,xu2024spot} point out, however, peer prediction has emphasized proving strategic alignment while devoting little attention to the statistical alignment of the resulting scores, limiting their direct use as evaluation metrics. Our test principles target both requirements jointly.

We list the direct optimization of evaluation metrics for alignment as future work. This connects to prior work on optimal scoring rule design \citep{li2022optimization, neyman2021binary, hartline2022optimal, chen2023learning, papireddygari2022contracts, chen2024optimal}, where the designer optimizes over scoring rules, with objectives such as informativeness, effort incentives, risk, or downstream decision quality, and subject to the constraint of truthfulness. In text evaluation, alignment with human judgments becomes the design objective, subject to robustness against strategic perturbations. \citet{lu2025aligned} take a step in this direction by optimizing aligned textual scoring rules with access to ground-truth texts. We leave the corresponding problem without ground truth, jointly optimizing for statistical and strategic alignment, to future work.

\section{Conclusion and Discussion}

We study reference-based text evaluation metrics. This perspective separates statistical alignment, measured by correlation with human ratings, from strategic alignment, measured by robustness to degradations and manipulations. We introduced unified test principles for these two requirements and a design framework for MI-based metrics. Empirically, LLM-as-a-Judge achieves strong human-rating correlation but is vulnerable to manipulation, while MI-based metrics, especially the statement-level TV $f$-variational metric, newly derived from our design framework, offer stronger strategic robustness while remaining competitive on statistical alignment.

Future work should optimize evaluation metrics directly for statistical and strategic alignment, as suggested in \Cref{sec: related work}.
Another direction is to reduce score variance. Our LLM-oracle metrics suffer randomness from oracle judgments, statement decomposition, and negative-reference sampling. This variance can obscure both human-rating correlation and robustness tests. Repeated oracle calls, multiple independent decompositions, larger negative-reference pools, and adaptive resampling near decision boundaries are simple ways to reduce estimation noise and make comparisons between mechanisms more reliable.

The evaluation framework itself can also be strengthened. Our current degradation and manipulation tests use fixed perturbation families. In deployment, however, models optimized against a metric may discover new score-inflating strategies. Future evaluation workflows should therefore include adaptive or competitive tests in which an adversary searches for perturbations that increase the score without adding task-relevant information. Such tests would evaluate a metric as an optimization target, not only as a static measurement tool.

Finally, reference-based alignment is limited by the reference signal. As language models exceed individual human references on some tasks, evaluation datasets may need richer reference sets, such as expert-panel references, debate- or critique-based references, or aggregated LLM outputs \citep{feng2026peer}. Strategically validated evaluation metrics may also serve as training rewards with reduced susceptibility to reward hacking. For example, \citet{feng2026peer} use peer-prediction-inspired self-training signals for language-model reasoning without relying on gold labels. This suggests a broader research program in which reference construction, evaluation-metric design, and
training objectives are studied jointly.

\begin{ack}
This work is supported by United States National Science Foundation Award \#2313137.
\end{ack}

\bibliographystyle{plainnat}

\newpage
\appendix

 \section{Design Framework Details}

\subsection{Direct Density Ratio Estimation}
\label{apdx: estimation direct density ratio}

For KL, $f(t)=t\log t$, so
\begin{equation*}
    I(X;Y)
    =
    \E_{\Pmarg}\bigl[r(X,Y)\log r(X,Y)\bigr]
    =
    \E_{\Pjoint}\bigl[\log r(X,Y)\bigr].
\end{equation*}
The quantity $\PMI(x; y) \coloneqq \log r(x,y) = \log p(y \mid x) - \log p(y)$
is the \emph{pointwise mutual information}; its expectation under the joint distribution recovers the Shannon mutual information. 

Direct estimation is also available for TVD-MI, where \(f(t)=\frac{1}{2}|t-1|\), so we have
\begin{equation*}
I_{\mathrm{TV}}(X;Y) = \frac{1}{2}\E_{\Pmarg}\bigl[|r(X,Y)-1|\bigr] = \frac{1}{2}\sum_{x,y\in \mathcal{V}^{*}}|P(x,y) - P(x)P(y)|.
\end{equation*}
This can also be written as $I_{\mathrm{TV}}(X;Y) = \frac{1}{2}\E_{\Pjoint}\bigl[|1-\frac{1}{r(X,Y)}|\bigr]$, providing two approaches of Monte Carlo estimation.

\subsection{$f$-Variational Estimation}
\label{apdx: estimation f variational}
Every $f$-divergence admits a variational lower bound via the Fenchel conjugate $f^*$.

\begin{definition}[Variational representation]\label{def:f-variational}
Let $f^*$ denote the Fenchel conjugate of $f$,
\[
f^*(v) \coloneqq \sup_{t>0}\{vt-f(t)\}.
\]
Then
\[
D_f(P\|Q)
=
\sup_{T}
\left\{
\E_P[T(X)]-\E_Q\bigl[f^*(T(X))\bigr]
\right\},
\]
where the supremum is over functions $T$ such that $f^*(T(x)) < \infty$ for all $x$ with $Q(x)>0$.
\end{definition}

Here $T$ is a critic: a scalar scoring function that tries to assign higher values to samples from $P$ than to samples from $Q$. We write $T^*$ for the population-optimal critic. When $f$ is differentiable,
\[
T^*(x) = f'\left(\frac{p(x)}{q(x)}\right),
\]
and in general
\[
T^*(x) \in \partial f\left(\frac{p(x)}{q(x)}\right),
\]
where $\partial f$ denotes the subdifferential of $f$.

The $f$-mutual information also admits the variational form
\[
\fMI(X;Y)
=
\sup_{T}
\left\{
\mathbb{E}_{P_{XY}}[T(X,Y)]
-
\mathbb{E}_{P_X\otimes P_Y}[f^*(T(X,Y))]
\right\}.
\]
At the population optimum, equality is attained when $T(x,y)=T^*(x,y)$.

In our setting, we can write $T(x,y)$ for a critic on candidate--reference pairs and $T^*(x,y)$ for the population-optimal critic. The optimal critic is determined by the chosen generator $f$ through
\[
T^*(x,y) \in \partial f\bigl(r(x,y)\bigr).
\]
As Table~\ref{tab:fmi-best-distinguishers} shows, different choices of $f$ induce different optimal critics and different conjugate penalty terms. We refer back to this table when instantiating KL-based and TV-based evaluation metrics.

\begin{table}[t]
\centering
\small
\caption{Common $f$-divergences, along with a convenient population-optimal critic $T^*(x,y)$ and the corresponding conjugate term $f^*(T^*(x,y))$, written in terms of the density ratio $r(x,y)$.}
\label{tab:fmi-best-distinguishers}
\begin{tabular}{@{}llll@{}}
\toprule
$f$-divergence & $f(t)$ & $T^*(x,y)$ & $f^*(T^*(x,y))$ \\
\midrule
Total variation
& $\frac{1}{2}|t-1|$
& $\frac{1}{2}\operatorname{sign}\bigl(r(x,y)-1\bigr)$
& $\frac{1}{2}\operatorname{sign}\bigl(r(x,y)-1\bigr)$ \\

KL divergence
& $t\log t$
& $1+\log r(x,y)$
& $r(x,y)$ \\

Reverse KL
& $-\log t$
& $-\frac{1}{r(x,y)}$
& $-1+\log r(x,y)$ \\

Pearson $\chi^2$
& $(t-1)^2$
& $2\bigl(r(x,y)-1\bigr)$
& $r(x,y)^2-1$ \\

Squared Hellinger
& $(\sqrt{t}-1)^2$
& $1-\frac{1}{\sqrt{r(x,y)}}$
& $\sqrt{r(x,y)}-1$ \\
\bottomrule
\end{tabular}
\end{table}

\paragraph{Estimator} Here joint samples $\Pjoint$ are candidate-reference pairs from the same task, while marginal samples $\Pmarg$ are pairs drawn across different tasks. Recall that, the optimum is attained by a critic satisfying
\(T^*(x,y)\in \partial f\bigl(r(x,y)\bigr)\). For common choices of $f$, the corresponding forms of $T^*(x,y)$ and $f^*(T^*(x,y))$ are listed in Table~\ref{tab:fmi-best-distinguishers}.

For KL, the variational estimation yields the NWJ objective:
\begin{align*}
I(X;Y) &= \sup_{T} \Bigl\{ \E_{\Pjoint}[T(X,Y)] - e^{-1} \E_{\Pmarg}\bigl[e^{T(X,Y)}\bigr] \Bigr\} \\
&= 1 + \E_{\Pjoint}\bigl[\log r(X,Y)\bigr] - \E_{\Pmarg}\bigl[r(X,Y)\bigr]
\end{align*}

For total variation, the estimator reduces to
\begin{align}
    I_{\mathrm{TV}}(X;Y) = & \sup_{\|T\|_\infty \le 1/2} \Bigl\{ \E_{\Pjoint}[T(X,Y)] - \E_{\Pmarg}[T(X,Y)] \Bigr\} \label{eq:tv-var}\\
    = & \frac{1}{2}\E_{\Pjoint}\bigl[\operatorname{sign}(r(X,Y)-1)\bigr]
         - \frac{1}{2}\E_{\Pmarg}\bigl[\operatorname{sign}(r(X,Y)-1)\bigr] \nonumber
\end{align}
where the optimal critic $\operatorname{sign}(r(X,Y)-1)$ is equivalent to a binary classifier that distinguishes same-task pairs from cross-task pairs.
 \section{TV, f-variational, Statement-level, LLM-oracle Metric}
\label{sec:contrastive-tv}

This method instantiates the scoring framework with
\[
F=\TV,\qquad E=\text{$f$-variational},\qquad D=\text{LLM-as-a-Judge (statement-level) denoted as }D_\Psi,
\]
where \(D_\Psi\) is induced by a reference decomposition \(\Psi\).

We keep the candidate at the report level and decompose each reference into
statements:
\begin{equation}
    \Psi(y)=\bigl(\psi_1(y),\ldots,\psi_{K(y)}(y)\bigr).
\end{equation}
We instantiate the local critic \(t\) and aggregation rule \(A\) from
Section~\ref{subsec:prediction} as follows. For each candidate--statement
pair, an LLM judge returns a centered binary local critic
\begin{equation}\label{eq:binary-align}
    t:\mathcal{V}^{*}\times \mathcal{V}^{*}\to \left\{-\frac12,\frac12\right\},
\end{equation}
where \(t(x,w)=\frac12\) if candidate \(x\) supports or aligns with statement
\(w\), and \(t(x,w)=-\frac12\) otherwise. We use mean aggregation, so the induced report-level critic is
\begin{equation}\label{eq:mean-align}
    T_{\Psi}(x,y)
    =
    \frac{1}{K(y)}
    \sum_{k=1}^{K(y)} t\bigl(x,\psi_k(y)\bigr).
\end{equation}
Since each summand lies in \([ -\tfrac12,\tfrac12 ]\), \(T_\Psi\) satisfies the constraint in the TV variational estimator. 

For task \(i\), let \(\mathcal{R}_i^+\) be positive references paired with the
same task, and let \(\mathcal{R}_i^-\) be negative references sampled from the
marginal reference distribution \(P_Y\). We define the sampled contrastive score
\begin{equation}\label{eq:contrastive-score}
    \hat S_{\mathrm{CTV}}(x_i;\mathcal{R}_i^+,\mathcal{R}_i^-)
    =
    \frac{1}{|\mathcal{R}_i^+|}
    \sum_{y\in\mathcal{R}_i^+} T_{\Psi}(x_i,y)
    -
    \frac{1}{|\mathcal{R}_i^-|}
    \sum_{y\in\mathcal{R}_i^-} T_{\Psi}(x_i,y).
\end{equation}

Assume that, conditional on task \(\psi_i\), the references in
\(\mathcal{R}_i^+\) are drawn i.i.d.\ from \(P_{Y\mid \psi_i}\), and the references in \(\mathcal{R}_i^-\) are drawn
i.i.d.\ from \(P_Y\), independently of \((x_i,\psi_i)\). Then
\begin{equation}\label{eq:contrastive-target-cond}
    \E\left[\hat S_{\mathrm{CTV}}(x_i;\mathcal{R}_i^+,\mathcal{R}_i^-)\mid x_i,\psi_i\right]
    =
    \E_{Y\sim P_{Y\mid \psi_i}}\bigl[T_\Psi(x_i,Y)\bigr]
    -
    \E_{Y\sim P_Y}\bigl[T_\Psi(x_i,Y)\bigr].
\end{equation}
If \((X_i,Y_i)\) is generated by first sampling a task and then sampling the candidate and reference independently conditional on that task, averaging Eq.~\eqref{eq:contrastive-target-cond} over \((X_i,\psi_i)\) yields
\begin{equation}\label{eq:contrastive-target}
    \E\bigl[\hat S_{\mathrm{CTV}}(X_i;\mathcal{R}_i^+,\mathcal{R}_i^-)\bigr]
    =
    \E_{\Pjoint}\bigl[T_\Psi(X,Y)\bigr]
    -
    \E_{\Pmarg}\bigl[T_\Psi(X,Y)\bigr].
\end{equation}
Because \(T_\Psi\) factors through \(\Psi(Y)\) and satisfies
\(\|T_\Psi\|_\infty\le 1/2\), Eq.~\eqref{eq:tv-var} applied to the transformed
pair \((X,\Psi(Y))\) gives
\[
    \E\bigl[\hat S_{\mathrm{CTV}}(X_i;\mathcal{R}_i^+,\mathcal{R}_i^-)\bigr]
    \le
    I_{\TV}\bigl(X;\Psi(Y)\bigr)
    \le
    I_{\TV}(X;Y).
\]
 \section{Experiment Details}

\subsection{Experiment Design}
\label{apdx: experiment design}

To compare scoring methods in a controlled way, we organize our experiments around the design space induced by Definition~\ref{def:triple}. Each scoring method corresponds to a choice along four axes:
\begin{enumerate}[label=(\roman*)]
    \item the information measure $F \in \{\KL, \TV\}$;
    \item the estimator $E \in \{\text{Direct}, \text{$f$-variational}\}$;
    \item the representation $R \in \{\text{Token}, \text{Whole-report}, \text{Statement}\}$.
    \item the predictor $\Pi \in \{\text{Autoregression}, \text{LLM-oracle}\}$;
\end{enumerate}

The predictor and representation do not vary independently: autoregression composes only with token-level representation, while the LLM-oracle composes with whole-report or statement-level representation (Section~\ref{subsec:prediction}). The joint $(\Pi, R)$ axis therefore has three valid settings, yielding a $2 \times 2 \times 3 = 12$-cell factorial design (Table~\ref{tab:design-coverage}). Of the 12 cells, three correspond to previously proposed methods (GEM, GPPM-Judgment, and TVD-MI). By holding three axes fixed and varying one, we can isolate the marginal effect of each design choice on alignment, degradation sensitivity, and manipulation resistance.

\paragraph{Implementation.}
We instantiate the token-autoregressive model with \texttt{Llama-3.1-8B}, and adopt the pre-processing process from \citet{lu2024eliciting,xu2025benchmarking} with \texttt{claude-haiku-4.5}. 

We instantiate the LLM-oracle with \texttt{claude-haiku-4.5}, queried through OpenRouter API at temperature $0$ with a single sample per prompt. In the $f$-variational settings, the oracle output is used directly as the critic $T(x,y)$. In the direct-estimation settings, the oracle returns a 7-point ordinal supporting judgment, which we map to
\(\{1/8,\,1/4,\,1/2,\,1,\,2,\,4,\,8\}\)
and treat as a discretized surrogate for the density ratio estimate $\hat r(x,y)$. For statement-level representation, we first decompose the reference $y$ into atomic claims using a dataset-specific decomposition prompt, and then score the resulting (candidate, reference statement) pairs with the same claude-haiku-4.5 model.

Unless otherwise noted, each candidate is scored against one positive reference, and, for $f$-variational metrics, four negative references sampled from other tasks in the same dataset group and takes average to reduce noise. A group consists of same-kind tasks within a dataset, such as the same class in Peer Grading-WH or the same topic category (e.g., math or technology) in LFQA-E. ICLR 2026, Peer Grading-XLSK, SummEval, and SPACE are treated as a single group.

\subsection{Computational Resources}

All experiments were conducted on a high-performance workstation with the following specifications:

\paragraph{Hardware.} The machine is equipped with dual Intel Xeon Platinum 8470Q CPUs (totaling 104 physical cores and 208 threads), 1.0 TiB of system RAM, and one NVIDIA RTX PRO 6000 (Blackwell architecture) GPU with 96 GB of device memory.

\paragraph{Local Inference.} For metrics using the Llama-3.1-8B model, local inference was performed with an average processing speed of approximately 0.5s/it. The total local compute time required for the entire study (excluding API latency) was approximately 6 GPU hours.

\paragraph{API Usage and Reliability.} Evaluations involving Claude-Haiku-4.5 were conducted via the OpenRouter API.
\begin{itemize}
    \item Configuration: All API calls used a default max token limit of 4000 and a temperature setting of 0 to ensure deterministic and complete responses.
    \item Retry Logic: To handle potential network instability or API timeouts, we implemented a robust retry mechanism with a maximum of 4 attempts per request; a call was marked as a failure only if all retries were exhausted.
    \item Error Handling: In cases where specific perturbation methods (for degradation or manipulation tests) encountered API errors or generation failures, the system was designed to gracefully fall back to a no-op (no operation), ensuring the stability of the overall evaluation pipeline.
\end{itemize}

\paragraph{Execution Time.} Due to varying dataset sizes, the total experimental duration per dataset ranged from 20 minutes to 1 hour. 
\section{Detailed Degradation and Manipulation Statistics}
\label{app:degradation-manipulation-statistics}

\begingroup
\scriptsize
\setlength{\tabcolsep}{2pt}
\renewcommand{\arraystretch}{0.92}
\begin{longtable}{@{}p{0.245\linewidth}*{7}{>{\raggedleft\arraybackslash}p{0.099\linewidth}}@{}}
\caption{Degradation perturbation method results. Each cell reports the standardized mean difference $d$ with 95\% CI. Red marks significant score increases ($p > 0.05$); orange marks non-significant changes. Both flag failed degradation criteria.} \\
\toprule
\textbf{Metric} & PG-WH & PG-XLSK & ICLR26 & SummEval & SPACE & LFQA-E & MedAESQA \\
\midrule
\endfirsthead

\caption[]{Degradation perturbation method results (continued). Each cell reports the standardized mean difference $d$ with 95\% CI. Red marks significant score increases ($p > 0.05$); orange marks non-significant changes. Both flag failed degradation criteria.} \\
\toprule
\textbf{Metric} & PG-WH & PG-XLSK & ICLR26 & SummEval & SPACE & LFQA-E & MedAESQA \\
\midrule
\endhead

\midrule
\multicolumn{8}{r}{\textit{Continued on next page}} \\
\endfoot

\bottomrule
\endlastfoot

\multicolumn{8}{@{}c@{}}{\textbf{\texttt{deletion\_and\_completion}}} \\
\specialrule{\lightrulewidth}{0.25em}{0.35em}
\multicolumn{8}{@{}l}{\textit{MI-based Mechanisms}} \\
KL-Direct-Autoreg. & $-0.45{\pm}0.09$ & $-0.32{\pm}0.08$ & $-0.24{\pm}0.07$ & $-0.41{\pm}0.09$ & $-0.50{\pm}0.11$ & $-0.10{\pm}0.09$ & $-0.21{\pm}0.09$ \\
TVD-FVar-Report & \textcolor{orange}{$-0.04{\pm}0.16$} & $-0.19{\pm}0.11$ & $-0.20{\pm}0.12$ & $-0.23{\pm}0.15$ & $-0.41{\pm}0.13$ & $-0.22{\pm}0.14$ & \textcolor{red}{$+0.19{\pm}0.14$} \\
TVD-FVar-Statement & $-0.20{\pm}0.10$ & $-0.23{\pm}0.11$ & $-0.18{\pm}0.12$ & $-0.48{\pm}0.10$ & $-0.31{\pm}0.09$ & $-0.19{\pm}0.09$ & $-0.19{\pm}0.10$ \\
\midrule
\multicolumn{8}{@{}l}{\textit{Non-MI baselines}} \\
ROUGE-L & \textcolor{orange}{$+0.06{\pm}0.09$} & \textcolor{red}{$+0.06{\pm}0.04$} & $-0.55{\pm}0.08$ & \textcolor{orange}{$-0.05{\pm}0.07$} & $-0.27{\pm}0.07$ & $-0.24{\pm}0.07$ & $-0.14{\pm}0.06$ \\
BLEU & \textcolor{red}{$+0.24{\pm}0.07$} & \textcolor{orange}{$+0.05{\pm}0.06$} & $-0.35{\pm}0.06$ & \textcolor{orange}{$+0.03{\pm}0.06$} & $-0.24{\pm}0.09$ & $-0.16{\pm}0.08$ & $-0.10{\pm}0.05$ \\
BERTScore & \textcolor{orange}{$+0.04{\pm}0.18$} & \textcolor{orange}{$+0.09{\pm}0.11$} & \textcolor{red}{$+0.56{\pm}0.09$} & \textcolor{orange}{$-0.05{\pm}0.10$} & $-0.32{\pm}0.09$ & \textcolor{orange}{$+0.04{\pm}0.08$} & \textcolor{orange}{$-0.04{\pm}0.06$} \\
LLM-J / Claude-Haiku-4.5 & $-0.35{\pm}0.11$ & $-0.19{\pm}0.09$ & $-0.27{\pm}0.11$ & $-0.29{\pm}0.14$ & $-0.44{\pm}0.13$ & $-0.37{\pm}0.13$ & \textcolor{orange}{$-0.01{\pm}0.09$} \\
LLM-J / Claude-Sonnet-4.5 & $-0.37{\pm}0.10$ & $-0.16{\pm}0.09$ & $-0.28{\pm}0.10$ & $-0.42{\pm}0.11$ & $-0.61{\pm}0.13$ & $-0.39{\pm}0.11$ & $-0.12{\pm}0.08$ \\
LLM-J / GPT-5-mini & $-0.24{\pm}0.10$ & $-0.13{\pm}0.08$ & $-0.32{\pm}0.12$ & $-0.29{\pm}0.12$ & $-0.52{\pm}0.12$ & $-0.17{\pm}0.11$ & \textcolor{orange}{$-0.09{\pm}0.09$} \\
LLM-J / GPT-4o-mini & \textcolor{orange}{$+0.06{\pm}0.11$} & \textcolor{red}{$+0.18{\pm}0.07$} & $-0.13{\pm}0.12$ & $-0.15{\pm}0.12$ & $-0.12{\pm}0.12$ & $-0.13{\pm}0.11$ & $-0.10{\pm}0.06$ \\

\addlinespace[0.7em]
\specialrule{\heavyrulewidth}{0pt}{0.35em}
\multicolumn{8}{@{}c@{}}{\textbf{\texttt{opinion\_flip}}} \\
\specialrule{\lightrulewidth}{0.25em}{0.35em}
\multicolumn{8}{@{}l}{\textit{MI-based Mechanisms}} \\
KL-Direct-Autoreg. & $-0.39{\pm}0.12$ & $-0.51{\pm}0.12$ & $-0.44{\pm}0.09$ & -- & -- & -- & -- \\
TVD-FVar-Report & $-0.21{\pm}0.16$ & \textcolor{orange}{$-0.01{\pm}0.11$} & \textcolor{orange}{$-0.05{\pm}0.14$} & -- & -- & -- & -- \\
TVD-FVar-Statement & $-0.84{\pm}0.19$ & $-0.27{\pm}0.17$ & $-0.28{\pm}0.15$ & -- & -- & -- & -- \\
\midrule
\multicolumn{8}{@{}l}{\textit{Non-MI baselines}} \\
ROUGE-L & \textcolor{red}{$+0.28{\pm}0.13$} & $-0.85{\pm}0.10$ & \textcolor{red}{$+0.13{\pm}0.09$} & -- & -- & -- & -- \\
BLEU & \textcolor{red}{$+0.63{\pm}0.11$} & $-0.64{\pm}0.12$ & $-0.16{\pm}0.12$ & -- & -- & -- & -- \\
BERTScore & \textcolor{orange}{$-0.06{\pm}0.19$} & $-0.35{\pm}0.14$ & \textcolor{red}{$+0.39{\pm}0.12$} & -- & -- & -- & -- \\
LLM-J / Claude-Haiku-4.5 & $-0.88{\pm}0.21$ & $-0.67{\pm}0.20$ & $-0.58{\pm}0.18$ & -- & -- & -- & -- \\
LLM-J / Claude-Sonnet-4.5 & $-0.99{\pm}0.21$ & $-0.70{\pm}0.18$ & $-0.68{\pm}0.17$ & -- & -- & -- & -- \\
LLM-J / GPT-5-mini & $-1.09{\pm}0.22$ & $-0.32{\pm}0.15$ & $-0.54{\pm}0.17$ & -- & -- & -- & -- \\
LLM-J / GPT-4o-mini & $-0.26{\pm}0.19$ & \textcolor{red}{$+0.22{\pm}0.19$} & \textcolor{orange}{$+0.06{\pm}0.17$} & -- & -- & -- & -- \\

\addlinespace[0.7em]
\specialrule{\heavyrulewidth}{0pt}{0.35em}
\multicolumn{8}{@{}c@{}}{\textbf{\texttt{random\_replacement}}} \\
\specialrule{\lightrulewidth}{0.25em}{0.35em}
\multicolumn{8}{@{}l}{\textit{MI-based Mechanisms}} \\
KL-Direct-Autoreg. & $-0.32{\pm}0.18$ & $-1.12{\pm}0.22$ & $-1.82{\pm}0.18$ & $-2.00{\pm}0.18$ & $-1.05{\pm}0.18$ & $-1.36{\pm}0.17$ & $-0.67{\pm}0.17$ \\
TVD-FVar-Report & $-0.48{\pm}0.18$ & $-1.88{\pm}0.23$ & $-8.98{\pm}0.20$ & $-2.60{\pm}0.21$ & $-1.42{\pm}0.20$ & $-2.93{\pm}0.19$ & $-3.00{\pm}0.21$ \\
TVD-FVar-Statement & $-0.81{\pm}0.18$ & $-0.92{\pm}0.19$ & $-1.62{\pm}0.18$ & $-2.10{\pm}0.20$ & $-1.01{\pm}0.18$ & $-1.96{\pm}0.18$ & $-2.61{\pm}0.19$ \\
\midrule
\multicolumn{8}{@{}l}{\textit{Non-MI baselines}} \\
ROUGE-L & \textcolor{orange}{$+0.01{\pm}0.18$} & $-0.34{\pm}0.21$ & $-1.15{\pm}0.16$ & $-2.06{\pm}0.18$ & $-1.03{\pm}0.19$ & $-1.18{\pm}0.17$ & $-1.95{\pm}0.18$ \\
BLEU & \textcolor{orange}{$+0.01{\pm}0.17$} & $-0.23{\pm}0.22$ & $-0.34{\pm}0.15$ & $-1.07{\pm}0.19$ & $-0.96{\pm}0.20$ & $-0.92{\pm}0.19$ & $-0.94{\pm}0.20$ \\
BERTScore & \textcolor{orange}{$+0.06{\pm}0.19$} & $-0.28{\pm}0.19$ & $-0.57{\pm}0.15$ & $-2.36{\pm}0.20$ & $-0.86{\pm}0.15$ & $-1.89{\pm}0.17$ & $-1.98{\pm}0.20$ \\
LLM-J / Claude-Haiku-4.5 & $-0.61{\pm}0.19$ & $-1.45{\pm}0.22$ & $-1.25{\pm}0.20$ & $-2.49{\pm}0.20$ & $-1.13{\pm}0.18$ & $-2.72{\pm}0.20$ & $-2.95{\pm}0.19$ \\
LLM-J / Claude-Sonnet-4.5 & $-0.66{\pm}0.19$ & $-1.67{\pm}0.22$ & $-3.99{\pm}0.20$ & $-3.18{\pm}0.20$ & $-1.22{\pm}0.19$ & $-2.81{\pm}0.20$ & $-2.95{\pm}0.20$ \\
LLM-J / GPT-5-mini & $-0.56{\pm}0.20$ & $-1.23{\pm}0.22$ & $-3.47{\pm}0.20$ & $-3.01{\pm}0.20$ & $-1.03{\pm}0.19$ & $-3.10{\pm}0.20$ & $-3.17{\pm}0.20$ \\
LLM-J / GPT-4o-mini & $-0.34{\pm}0.20$ & \textcolor{orange}{$-0.11{\pm}0.23$} & \textcolor{orange}{$+0.08{\pm}0.20$} & $-3.74{\pm}0.20$ & $-0.64{\pm}0.18$ & $-3.10{\pm}0.19$ & $-2.49{\pm}0.19$ \\

\addlinespace[0.7em]
\specialrule{\heavyrulewidth}{0pt}{0.35em}
\multicolumn{8}{@{}c@{}}{\textbf{\texttt{sentence\_deletion}}} \\
\specialrule{\lightrulewidth}{0.25em}{0.35em}
\multicolumn{8}{@{}l}{\textit{MI-based Mechanisms}} \\
KL-Direct-Autoreg. & $-0.46{\pm}0.09$ & $-0.64{\pm}0.08$ & $-0.70{\pm}0.07$ & $-0.34{\pm}0.10$ & $-0.56{\pm}0.11$ & $-0.25{\pm}0.09$ & $-0.24{\pm}0.10$ \\
TVD-FVar-Report & \textcolor{orange}{$-0.05{\pm}0.16$} & $-0.20{\pm}0.11$ & $-0.21{\pm}0.15$ & $-0.16{\pm}0.15$ & $-0.31{\pm}0.12$ & $-0.23{\pm}0.14$ & \textcolor{orange}{$-0.03{\pm}0.15$} \\
TVD-FVar-Statement & $-0.25{\pm}0.11$ & $-0.22{\pm}0.11$ & $-0.38{\pm}0.10$ & $-0.50{\pm}0.10$ & $-0.27{\pm}0.08$ & $-0.18{\pm}0.09$ & $-0.45{\pm}0.11$ \\
\midrule
\multicolumn{8}{@{}l}{\textit{Non-MI baselines}} \\
ROUGE-L & \textcolor{orange}{$-0.04{\pm}0.08$} & \textcolor{red}{$+0.43{\pm}0.08$} & $-0.25{\pm}0.10$ & \textcolor{orange}{$-0.02{\pm}0.08$} & $-0.10{\pm}0.08$ & $-0.44{\pm}0.10$ & $-0.10{\pm}0.08$ \\
BLEU & $-0.19{\pm}0.11$ & \textcolor{red}{$+0.39{\pm}0.12$} & $-0.67{\pm}0.13$ & \textcolor{orange}{$+0.01{\pm}0.07$} & $-0.22{\pm}0.12$ & $-0.56{\pm}0.12$ & $-0.27{\pm}0.10$ \\
BERTScore & \textcolor{orange}{$+0.05{\pm}0.19$} & $-0.16{\pm}0.10$ & \textcolor{red}{$+0.12{\pm}0.10$} & $-0.17{\pm}0.10$ & $-0.16{\pm}0.08$ & \textcolor{orange}{$-0.04{\pm}0.08$} & $-0.11{\pm}0.06$ \\
LLM-J / Claude-Haiku-4.5 & $-0.39{\pm}0.13$ & $-0.71{\pm}0.11$ & $-0.88{\pm}0.15$ & $-0.37{\pm}0.14$ & $-0.53{\pm}0.13$ & $-0.86{\pm}0.14$ & $-0.34{\pm}0.11$ \\
LLM-J / Claude-Sonnet-4.5 & $-0.43{\pm}0.11$ & $-0.80{\pm}0.09$ & $-1.40{\pm}0.14$ & $-0.42{\pm}0.12$ & $-0.61{\pm}0.14$ & $-0.91{\pm}0.14$ & $-0.37{\pm}0.10$ \\
LLM-J / GPT-5-mini & $-0.43{\pm}0.10$ & $-0.73{\pm}0.08$ & $-1.11{\pm}0.14$ & $-0.44{\pm}0.12$ & $-0.63{\pm}0.12$ & $-0.70{\pm}0.13$ & $-0.41{\pm}0.10$ \\
LLM-J / GPT-4o-mini & $-0.59{\pm}0.12$ & $-0.73{\pm}0.09$ & $-1.04{\pm}0.16$ & $-0.54{\pm}0.13$ & $-0.58{\pm}0.14$ & $-1.01{\pm}0.15$ & $-0.60{\pm}0.10$ \\

\addlinespace[0.7em]
\specialrule{\heavyrulewidth}{0pt}{0.35em}
\multicolumn{8}{@{}c@{}}{\textbf{\texttt{surface\_report}}} \\
\specialrule{\lightrulewidth}{0.25em}{0.35em}
\multicolumn{8}{@{}l}{\textit{MI-based Mechanisms}} \\
KL-Direct-Autoreg. & -- & $-0.74{\pm}0.20$ & \textcolor{orange}{$-0.02{\pm}0.14$} & $-0.28{\pm}0.15$ & $-1.22{\pm}0.17$ & \textcolor{orange}{$+0.04{\pm}0.11$} & \textcolor{orange}{$-0.08{\pm}0.13$} \\
TVD-FVar-Report & -- & $-1.36{\pm}0.22$ & \textcolor{orange}{$-0.12{\pm}0.21$} & \textcolor{orange}{$-0.09{\pm}0.16$} & $-1.02{\pm}0.18$ & \textcolor{orange}{$+0.02{\pm}0.17$} & \textcolor{orange}{$-0.02{\pm}0.19$} \\
TVD-FVar-Statement & -- & $-0.58{\pm}0.17$ & $-0.31{\pm}0.18$ & \textcolor{orange}{$+0.04{\pm}0.15$} & $-0.84{\pm}0.18$ & \textcolor{orange}{$+0.11{\pm}0.12$} & \textcolor{orange}{$-0.01{\pm}0.18$} \\
\midrule
\multicolumn{8}{@{}l}{\textit{Non-MI baselines}} \\
ROUGE-L & -- & $-3.32{\pm}0.21$ & $-0.93{\pm}0.15$ & \textcolor{orange}{$-0.11{\pm}0.12$} & $-2.46{\pm}0.20$ & $-0.14{\pm}0.11$ & $-0.45{\pm}0.16$ \\
BLEU & -- & $-2.84{\pm}0.21$ & $-1.28{\pm}0.15$ & \textcolor{red}{$+0.68{\pm}0.16$} & $-1.34{\pm}0.20$ & \textcolor{orange}{$-0.03{\pm}0.13$} & $-0.56{\pm}0.17$ \\
BERTScore & -- & \textcolor{orange}{$+0.10{\pm}0.18$} & \textcolor{red}{$+0.84{\pm}0.13$} & \textcolor{red}{$+0.28{\pm}0.14$} & $-2.81{\pm}0.17$ & \textcolor{orange}{$-0.03{\pm}0.13$} & \textcolor{orange}{$-0.11{\pm}0.16$} \\
LLM-J / Claude-Haiku-4.5 & -- & $-1.57{\pm}0.22$ & \textcolor{red}{$+0.18{\pm}0.17$} & \textcolor{red}{$+0.47{\pm}0.17$} & $-1.81{\pm}0.19$ & \textcolor{red}{$+0.81{\pm}0.16$} & \textcolor{red}{$+1.21{\pm}0.19$} \\
LLM-J / Claude-Sonnet-4.5 & -- & $-1.65{\pm}0.22$ & \textcolor{orange}{$+0.05{\pm}0.18$} & \textcolor{red}{$+0.28{\pm}0.15$} & $-2.51{\pm}0.20$ & \textcolor{red}{$+0.69{\pm}0.16$} & \textcolor{red}{$+1.18{\pm}0.20$} \\
LLM-J / GPT-5-mini & -- & $-1.52{\pm}0.22$ & \textcolor{orange}{$+0.03{\pm}0.17$} & \textcolor{orange}{$+0.12{\pm}0.17$} & $-1.28{\pm}0.19$ & \textcolor{red}{$+0.51{\pm}0.14$} & \textcolor{red}{$+1.01{\pm}0.19$} \\
LLM-J / GPT-4o-mini & -- & $-1.11{\pm}0.22$ & \textcolor{red}{$+1.13{\pm}0.17$} & \textcolor{red}{$+1.39{\pm}0.17$} & $-0.74{\pm}0.18$ & \textcolor{red}{$+1.04{\pm}0.17$} & \textcolor{red}{$+0.98{\pm}0.19$} \\

\addlinespace[0.7em]
\specialrule{\heavyrulewidth}{0pt}{0.35em}
\multicolumn{8}{@{}c@{}}{\textbf{\texttt{ultra\_concise\_compression}}} \\
\specialrule{\lightrulewidth}{0.25em}{0.35em}
\multicolumn{8}{@{}l}{\textit{MI-based Mechanisms}} \\
KL-Direct-Autoreg. & $-0.61{\pm}0.12$ & $-1.16{\pm}0.12$ & $-1.43{\pm}0.13$ & $-0.89{\pm}0.15$ & $-1.31{\pm}0.17$ & \textcolor{orange}{$-0.10{\pm}0.11$} & \textcolor{orange}{$+0.01{\pm}0.13$} \\
TVD-FVar-Report & \textcolor{orange}{$-0.12{\pm}0.18$} & $-0.59{\pm}0.15$ & \textcolor{orange}{$+0.01{\pm}0.19$} & \textcolor{red}{$+0.19{\pm}0.15$} & $-0.52{\pm}0.15$ & \textcolor{orange}{$-0.03{\pm}0.16$} & \textcolor{red}{$+0.34{\pm}0.17$} \\
TVD-FVar-Statement & $-0.24{\pm}0.13$ & $-0.28{\pm}0.13$ & $-0.29{\pm}0.14$ & $-1.10{\pm}0.14$ & $-0.59{\pm}0.13$ & $-0.27{\pm}0.11$ & $-0.20{\pm}0.14$ \\
\midrule
\multicolumn{8}{@{}l}{\textit{Non-MI baselines}} \\
ROUGE-L & $-2.43{\pm}0.17$ & $-4.80{\pm}0.25$ & $-3.68{\pm}0.18$ & $-0.90{\pm}0.15$ & $-1.42{\pm}0.17$ & $-3.06{\pm}0.16$ & $-1.77{\pm}0.16$ \\
BLEU & $-2.18{\pm}0.20$ & $-3.24{\pm}0.23$ & $-3.78{\pm}0.20$ & $-1.26{\pm}0.20$ & $-1.25{\pm}0.20$ & $-1.77{\pm}0.20$ & $-1.47{\pm}0.20$ \\
BERTScore & \textcolor{red}{$+0.19{\pm}0.19$} & \textcolor{red}{$+0.81{\pm}0.15$} & \textcolor{red}{$+1.33{\pm}0.11$} & \textcolor{red}{$+0.19{\pm}0.14$} & $-0.33{\pm}0.14$ & \textcolor{orange}{$+0.03{\pm}0.11$} & \textcolor{orange}{$-0.06{\pm}0.11$} \\
LLM-J / Claude-Haiku-4.5 & $-0.33{\pm}0.11$ & $-0.96{\pm}0.13$ & $-0.37{\pm}0.13$ & $-0.40{\pm}0.16$ & $-0.75{\pm}0.15$ & $-0.90{\pm}0.16$ & $-0.76{\pm}0.14$ \\
LLM-J / Claude-Sonnet-4.5 & $-0.39{\pm}0.11$ & $-1.02{\pm}0.12$ & $-0.96{\pm}0.14$ & $-0.79{\pm}0.14$ & $-0.88{\pm}0.14$ & $-1.03{\pm}0.16$ & $-0.63{\pm}0.13$ \\
LLM-J / GPT-5-mini & $-0.18{\pm}0.10$ & $-0.88{\pm}0.11$ & $-0.99{\pm}0.13$ & $-0.54{\pm}0.13$ & $-0.84{\pm}0.14$ & $-0.79{\pm}0.13$ & $-0.62{\pm}0.14$ \\
LLM-J / GPT-4o-mini & $-0.85{\pm}0.14$ & $-1.03{\pm}0.11$ & $-1.02{\pm}0.18$ & $-1.42{\pm}0.18$ & $-1.15{\pm}0.16$ & $-1.70{\pm}0.16$ & $-1.40{\pm}0.12$ \\

\end{longtable}
\endgroup

\begingroup
\scriptsize
\setlength{\tabcolsep}{2pt}
\renewcommand{\arraystretch}{0.92}
\begin{longtable}{@{}p{0.245\linewidth}*{7}{>{\raggedleft\arraybackslash}p{0.099\linewidth}}@{}}
\caption{Manipulation perturbation method results. Each cell reports the standardized mean difference $d$ with 95\% CI. Red marks significant score increases ($p > 0.05$), which flags failed manipulation criteria.} \\
\toprule
\textbf{Metric} & PG-WH & PG-XLSK & ICLR26 & SummEval & SPACE & LFQA-E & MedAESQA \\
\midrule
\endfirsthead

\caption[]{Manipulation perturbation method results (continued). Each cell reports the standardized mean difference $d$ with 95\% CI. Red marks significant score increases ($p > 0.05$), which flags failed manipulation criteria.} \\
\toprule
\textbf{Metric} & PG-WH & PG-XLSK & ICLR26 & SummEval & SPACE & LFQA-E & MedAESQA \\
\midrule
\endhead

\midrule
\multicolumn{8}{r}{\textit{Continued on next page}} \\
\endfoot

\bottomrule
\endlastfoot

\multicolumn{8}{@{}c@{}}{\textbf{\texttt{meaningless\_elongation}}} \\
\specialrule{\lightrulewidth}{0.25em}{0.35em}
\multicolumn{8}{@{}l}{\textit{MI-based Mechanisms}} \\
KL-Direct-Autoreg. & \textcolor{red}{$+0.06{\pm}0.05$} & $-0.01{\pm}0.05$ & $+0.00{\pm}0.02$ & $-0.01{\pm}0.03$ & $+0.01{\pm}0.04$ & $+0.00{\pm}0.05$ & $-0.05{\pm}0.04$ \\
TVD-FVar-Report & $+0.05{\pm}0.14$ & $-0.03{\pm}0.09$ & $-0.02{\pm}0.13$ & $+0.12{\pm}0.13$ & $-0.16{\pm}0.11$ & $-0.04{\pm}0.13$ & $+0.05{\pm}0.12$ \\
TVD-FVar-Statement & $+0.05{\pm}0.06$ & $-0.03{\pm}0.08$ & $-0.03{\pm}0.07$ & $+0.01{\pm}0.05$ & $+0.02{\pm}0.06$ & $+0.05{\pm}0.07$ & $-0.03{\pm}0.05$ \\
\midrule
\multicolumn{8}{@{}l}{\textit{Non-MI baselines}} \\
ROUGE-L & \textcolor{red}{$+0.48{\pm}0.14$} & $-1.59{\pm}0.15$ & $-0.33{\pm}0.02$ & $-0.62{\pm}0.07$ & $-0.84{\pm}0.09$ & $-0.08{\pm}0.03$ & $-0.13{\pm}0.03$ \\
BLEU & \textcolor{red}{$+0.41{\pm}0.16$} & $-1.44{\pm}0.17$ & $-0.01{\pm}0.04$ & $-0.53{\pm}0.08$ & $-0.39{\pm}0.08$ & \textcolor{red}{$+0.03{\pm}0.02$} & $+0.03{\pm}0.03$ \\
BERTScore & $+0.10{\pm}0.18$ & $-0.12{\pm}0.08$ & $+0.00{\pm}0.04$ & $-0.02{\pm}0.04$ & $+0.03{\pm}0.06$ & $+0.00{\pm}0.04$ & $-0.01{\pm}0.03$ \\
LLM-J / Claude-Haiku-4.5 & \textcolor{red}{$+0.15{\pm}0.08$} & $-0.28{\pm}0.08$ & $-0.02{\pm}0.07$ & $+0.05{\pm}0.09$ & $-0.37{\pm}0.10$ & $-0.05{\pm}0.06$ & $-0.11{\pm}0.05$ \\
LLM-J / Claude-Sonnet-4.5 & \textcolor{red}{$+0.10{\pm}0.06$} & $-0.77{\pm}0.09$ & \textcolor{red}{$+0.14{\pm}0.06$} & $+0.04{\pm}0.08$ & $-0.52{\pm}0.11$ & $-0.08{\pm}0.06$ & $-0.09{\pm}0.04$ \\
LLM-J / GPT-5-mini & $+0.04{\pm}0.07$ & \textcolor{red}{$+0.10{\pm}0.06$} & $+0.05{\pm}0.10$ & $-0.02{\pm}0.08$ & $-0.05{\pm}0.09$ & $+0.06{\pm}0.09$ & $-0.01{\pm}0.08$ \\
LLM-J / GPT-4o-mini & \textcolor{red}{$+0.30{\pm}0.11$} & \textcolor{red}{$+0.43{\pm}0.09$} & \textcolor{red}{$+0.24{\pm}0.09$} & $-0.25{\pm}0.13$ & $-0.37{\pm}0.13$ & $+0.00{\pm}0.07$ & $-0.04{\pm}0.05$ \\

\addlinespace[0.7em]
\specialrule{\heavyrulewidth}{0pt}{0.35em}
\multicolumn{8}{@{}c@{}}{\textbf{\texttt{opinion\_shift\_negative}}} \\
\specialrule{\lightrulewidth}{0.25em}{0.35em}
\multicolumn{8}{@{}l}{\textit{MI-based Mechanisms}} \\
KL-Direct-Autoreg. & $+0.01{\pm}0.10$ & $-0.03{\pm}0.10$ & \textcolor{red}{$+0.29{\pm}0.08$} & -- & -- & -- & -- \\
TVD-FVar-Report & $+0.03{\pm}0.15$ & $-0.05{\pm}0.10$ & $-0.03{\pm}0.13$ & -- & -- & -- & -- \\
TVD-FVar-Statement & $-0.19{\pm}0.14$ & $-0.20{\pm}0.13$ & $-0.20{\pm}0.10$ & -- & -- & -- & -- \\
\midrule
\multicolumn{8}{@{}l}{\textit{Non-MI baselines}} \\
ROUGE-L & $-0.55{\pm}0.17$ & $-3.34{\pm}0.22$ & $-1.10{\pm}0.11$ & -- & -- & -- & -- \\
BLEU & $-0.12{\pm}0.19$ & $-2.90{\pm}0.22$ & $-1.00{\pm}0.15$ & -- & -- & -- & -- \\
BERTScore & $-0.06{\pm}0.02$ & $-0.25{\pm}0.16$ & \textcolor{red}{$+0.42{\pm}0.11$} & -- & -- & -- & -- \\
LLM-J / Claude-Haiku-4.5 & \textcolor{red}{$+0.27{\pm}0.13$} & $-0.25{\pm}0.18$ & $+0.07{\pm}0.15$ & -- & -- & -- & -- \\
LLM-J / Claude-Sonnet-4.5 & \textcolor{red}{$+0.16{\pm}0.13$} & $-0.35{\pm}0.18$ & $+0.12{\pm}0.15$ & -- & -- & -- & -- \\
LLM-J / GPT-5-mini & \textcolor{red}{$+0.34{\pm}0.14$} & \textcolor{red}{$+0.34{\pm}0.15$} & $+0.10{\pm}0.14$ & -- & -- & -- & -- \\
LLM-J / GPT-4o-mini & \textcolor{red}{$+0.50{\pm}0.16$} & \textcolor{red}{$+0.24{\pm}0.17$} & \textcolor{red}{$+0.63{\pm}0.15$} & -- & -- & -- & -- \\

\addlinespace[0.7em]
\specialrule{\heavyrulewidth}{0pt}{0.35em}
\multicolumn{8}{@{}c@{}}{\textbf{\texttt{opinion\_shift\_neutral}}} \\
\specialrule{\lightrulewidth}{0.25em}{0.35em}
\multicolumn{8}{@{}l}{\textit{MI-based Mechanisms}} \\
KL-Direct-Autoreg. & $-0.23{\pm}0.08$ & $-0.03{\pm}0.07$ & \textcolor{red}{$+0.13{\pm}0.06$} & -- & -- & $-0.01{\pm}0.09$ & $-0.09{\pm}0.08$ \\
TVD-FVar-Report & $+0.05{\pm}0.15$ & $+0.01{\pm}0.10$ & $-0.05{\pm}0.16$ & -- & -- & $+0.00{\pm}0.13$ & $-0.04{\pm}0.15$ \\
TVD-FVar-Statement & $-0.04{\pm}0.08$ & $-0.13{\pm}0.10$ & $-0.19{\pm}0.08$ & -- & -- & $+0.00{\pm}0.09$ & $-0.08{\pm}0.07$ \\
\midrule
\multicolumn{8}{@{}l}{\textit{Non-MI baselines}} \\
ROUGE-L & \textcolor{red}{$+0.19{\pm}0.14$} & $-1.42{\pm}0.18$ & $-1.20{\pm}0.12$ & -- & -- & $-0.20{\pm}0.09$ & $-0.19{\pm}0.09$ \\
BLEU & \textcolor{red}{$+0.58{\pm}0.14$} & $-1.83{\pm}0.20$ & $-2.14{\pm}0.17$ & -- & -- & $-0.10{\pm}0.10$ & $-0.09{\pm}0.07$ \\
BERTScore & $-0.02{\pm}0.18$ & $-0.60{\pm}0.13$ & \textcolor{red}{$+0.18{\pm}0.10$} & -- & -- & $-0.01{\pm}0.09$ & $-0.11{\pm}0.05$ \\
LLM-J / Claude-Haiku-4.5 & \textcolor{red}{$+0.16{\pm}0.08$} & \textcolor{red}{$+0.56{\pm}0.12$} & \textcolor{red}{$+0.59{\pm}0.13$} & -- & -- & \textcolor{red}{$+0.48{\pm}0.13$} & \textcolor{red}{$+0.27{\pm}0.08$} \\
LLM-J / Claude-Sonnet-4.5 & \textcolor{red}{$+0.13{\pm}0.08$} & \textcolor{red}{$+0.44{\pm}0.14$} & \textcolor{red}{$+0.54{\pm}0.11$} & -- & -- & \textcolor{red}{$+0.30{\pm}0.13$} & \textcolor{red}{$+0.10{\pm}0.07$} \\
LLM-J / GPT-5-mini & \textcolor{red}{$+0.12{\pm}0.08$} & \textcolor{red}{$+0.52{\pm}0.09$} & \textcolor{red}{$+0.40{\pm}0.12$} & -- & -- & \textcolor{red}{$+0.39{\pm}0.13$} & \textcolor{red}{$+0.11{\pm}0.07$} \\
LLM-J / GPT-4o-mini & \textcolor{red}{$+0.74{\pm}0.12$} & \textcolor{red}{$+1.35{\pm}0.15$} & \textcolor{red}{$+1.09{\pm}0.17$} & -- & -- & \textcolor{red}{$+0.79{\pm}0.15$} & \textcolor{red}{$+0.17{\pm}0.07$} \\

\addlinespace[0.7em]
\specialrule{\heavyrulewidth}{0pt}{0.35em}
\multicolumn{8}{@{}c@{}}{\textbf{\texttt{opinion\_shift\_positive}}} \\
\specialrule{\lightrulewidth}{0.25em}{0.35em}
\multicolumn{8}{@{}l}{\textit{MI-based Mechanisms}} \\
KL-Direct-Autoreg. & $-0.41{\pm}0.08$ & $-0.15{\pm}0.07$ & \textcolor{red}{$+0.13{\pm}0.07$} & -- & -- & -- & -- \\
TVD-FVar-Report & $-0.06{\pm}0.14$ & $-0.01{\pm}0.10$ & $-0.07{\pm}0.13$ & -- & -- & -- & -- \\
TVD-FVar-Statement & $-0.14{\pm}0.10$ & $-0.13{\pm}0.10$ & $+0.02{\pm}0.09$ & -- & -- & -- & -- \\
\midrule
\multicolumn{8}{@{}l}{\textit{Non-MI baselines}} \\
ROUGE-L & $-0.29{\pm}0.16$ & $-2.04{\pm}0.20$ & $-0.59{\pm}0.08$ & -- & -- & -- & -- \\
BLEU & \textcolor{red}{$+0.19{\pm}0.18$} & $-2.00{\pm}0.21$ & $-0.46{\pm}0.11$ & -- & -- & -- & -- \\
BERTScore & $-0.13{\pm}0.02$ & $-0.56{\pm}0.13$ & $+0.03{\pm}0.09$ & -- & -- & -- & -- \\
LLM-J / Claude-Haiku-4.5 & $-0.15{\pm}0.10$ & \textcolor{red}{$+0.38{\pm}0.12$} & $+0.01{\pm}0.15$ & -- & -- & -- & -- \\
LLM-J / Claude-Sonnet-4.5 & $-0.19{\pm}0.11$ & \textcolor{red}{$+0.32{\pm}0.14$} & \textcolor{red}{$+0.18{\pm}0.12$} & -- & -- & -- & -- \\
LLM-J / GPT-5-mini & $+0.00{\pm}0.10$ & \textcolor{red}{$+0.54{\pm}0.09$} & \textcolor{red}{$+0.20{\pm}0.12$} & -- & -- & -- & -- \\
LLM-J / GPT-4o-mini & \textcolor{red}{$+1.09{\pm}0.13$} & \textcolor{red}{$+1.74{\pm}0.15$} & \textcolor{red}{$+1.23{\pm}0.18$} & -- & -- & -- & -- \\

\addlinespace[0.7em]
\specialrule{\heavyrulewidth}{0pt}{0.35em}
\multicolumn{8}{@{}c@{}}{\textbf{\texttt{opinion\_shift\_strong}}} \\
\specialrule{\lightrulewidth}{0.25em}{0.35em}
\multicolumn{8}{@{}l}{\textit{MI-based Mechanisms}} \\
KL-Direct-Autoreg. & $+0.01{\pm}0.07$ & \textcolor{red}{$+0.12{\pm}0.08$} & \textcolor{red}{$+0.14{\pm}0.05$} & -- & -- & $+0.04{\pm}0.09$ & $-0.09{\pm}0.09$ \\
TVD-FVar-Report & $+0.01{\pm}0.14$ & $-0.04{\pm}0.11$ & $-0.05{\pm}0.14$ & -- & -- & $-0.15{\pm}0.14$ & $+0.02{\pm}0.13$ \\
TVD-FVar-Statement & $+0.00{\pm}0.07$ & $-0.04{\pm}0.10$ & $-0.04{\pm}0.08$ & -- & -- & $-0.01{\pm}0.09$ & $+0.01{\pm}0.07$ \\
\midrule
\multicolumn{8}{@{}l}{\textit{Non-MI baselines}} \\
ROUGE-L & \textcolor{red}{$+0.45{\pm}0.12$} & $-2.91{\pm}0.19$ & $-1.18{\pm}0.10$ & -- & -- & $-0.25{\pm}0.09$ & $-0.07{\pm}0.09$ \\
BLEU & \textcolor{red}{$+0.64{\pm}0.13$} & $-2.81{\pm}0.21$ & $-2.21{\pm}0.16$ & -- & -- & $-0.16{\pm}0.12$ & $-0.13{\pm}0.07$ \\
BERTScore & $+0.00{\pm}0.02$ & $-0.39{\pm}0.14$ & \textcolor{red}{$+0.25{\pm}0.10$} & -- & -- & $+0.01{\pm}0.09$ & $-0.07{\pm}0.05$ \\
LLM-J / Claude-Haiku-4.5 & \textcolor{red}{$+0.14{\pm}0.09$} & \textcolor{red}{$+0.30{\pm}0.13$} & \textcolor{red}{$+0.49{\pm}0.13$} & -- & -- & \textcolor{red}{$+0.40{\pm}0.12$} & \textcolor{red}{$+0.24{\pm}0.08$} \\
LLM-J / Claude-Sonnet-4.5 & \textcolor{red}{$+0.13{\pm}0.09$} & \textcolor{red}{$+0.30{\pm}0.18$} & \textcolor{red}{$+0.47{\pm}0.11$} & -- & -- & \textcolor{red}{$+0.18{\pm}0.13$} & $+0.05{\pm}0.07$ \\
LLM-J / GPT-5-mini & \textcolor{red}{$+0.11{\pm}0.08$} & \textcolor{red}{$+0.56{\pm}0.11$} & \textcolor{red}{$+0.32{\pm}0.11$} & -- & -- & \textcolor{red}{$+0.24{\pm}0.11$} & \textcolor{red}{$+0.09{\pm}0.09$} \\
LLM-J / GPT-4o-mini & \textcolor{red}{$+0.52{\pm}0.11$} & \textcolor{red}{$+1.26{\pm}0.15$} & \textcolor{red}{$+1.08{\pm}0.17$} & -- & -- & \textcolor{red}{$+0.62{\pm}0.15$} & \textcolor{red}{$+0.14{\pm}0.07$} \\

\addlinespace[0.7em]
\specialrule{\heavyrulewidth}{0pt}{0.35em}
\multicolumn{8}{@{}c@{}}{\textbf{\texttt{rephrase}}} \\
\specialrule{\lightrulewidth}{0.25em}{0.35em}
\multicolumn{8}{@{}l}{\textit{MI-based Mechanisms}} \\
KL-Direct-Autoreg. & $-0.06{\pm}0.07$ & $-0.01{\pm}0.06$ & \textcolor{red}{$+0.07{\pm}0.05$} & $-0.09{\pm}0.06$ & $-0.35{\pm}0.08$ & $+0.01{\pm}0.08$ & $-0.15{\pm}0.09$ \\
TVD-FVar-Report & $-0.01{\pm}0.16$ & $-0.02{\pm}0.10$ & $-0.11{\pm}0.13$ & $-0.11{\pm}0.14$ & $-0.07{\pm}0.11$ & $-0.16{\pm}0.13$ & $+0.08{\pm}0.15$ \\
TVD-FVar-Statement & $+0.00{\pm}0.07$ & $-0.12{\pm}0.09$ & $-0.14{\pm}0.08$ & $-0.04{\pm}0.06$ & $-0.02{\pm}0.07$ & $-0.04{\pm}0.08$ & $-0.08{\pm}0.06$ \\
\midrule
\multicolumn{8}{@{}l}{\textit{Non-MI baselines}} \\
ROUGE-L & \textcolor{red}{$+0.47{\pm}0.13$} & $-1.97{\pm}0.14$ & $-0.98{\pm}0.09$ & $-0.09{\pm}0.07$ & $-0.62{\pm}0.12$ & $-0.16{\pm}0.07$ & $-0.21{\pm}0.09$ \\
BLEU & \textcolor{red}{$+0.68{\pm}0.13$} & $-2.53{\pm}0.20$ & $-2.09{\pm}0.15$ & \textcolor{red}{$+0.54{\pm}0.14$} & $-0.65{\pm}0.14$ & $-0.11{\pm}0.09$ & $-0.16{\pm}0.07$ \\
BERTScore & $-0.02{\pm}0.02$ & $-0.18{\pm}0.10$ & $+0.09{\pm}0.09$ & \textcolor{red}{$+0.13{\pm}0.09$} & $-0.58{\pm}0.12$ & $+0.02{\pm}0.08$ & $-0.16{\pm}0.06$ \\
LLM-J / Claude-Haiku-4.5 & \textcolor{red}{$+0.14{\pm}0.09$} & \textcolor{red}{$+0.18{\pm}0.09$} & \textcolor{red}{$+0.49{\pm}0.11$} & \textcolor{red}{$+0.29{\pm}0.10$} & \textcolor{red}{$+0.14{\pm}0.08$} & \textcolor{red}{$+0.25{\pm}0.11$} & \textcolor{red}{$+0.24{\pm}0.07$} \\
LLM-J / Claude-Sonnet-4.5 & \textcolor{red}{$+0.13{\pm}0.08$} & \textcolor{red}{$+0.19{\pm}0.09$} & \textcolor{red}{$+0.25{\pm}0.09$} & \textcolor{red}{$+0.28{\pm}0.09$} & $+0.00{\pm}0.09$ & $+0.01{\pm}0.12$ & $+0.02{\pm}0.07$ \\
LLM-J / GPT-5-mini & \textcolor{red}{$+0.09{\pm}0.08$} & \textcolor{red}{$+0.16{\pm}0.07$} & \textcolor{red}{$+0.17{\pm}0.10$} & $+0.07{\pm}0.08$ & $+0.01{\pm}0.08$ & \textcolor{red}{$+0.16{\pm}0.11$} & $+0.05{\pm}0.08$ \\
LLM-J / GPT-4o-mini & \textcolor{red}{$+0.52{\pm}0.11$} & \textcolor{red}{$+0.82{\pm}0.10$} & \textcolor{red}{$+0.81{\pm}0.15$} & \textcolor{red}{$+0.65{\pm}0.12$} & \textcolor{red}{$+0.36{\pm}0.11$} & \textcolor{red}{$+0.40{\pm}0.13$} & \textcolor{red}{$+0.21{\pm}0.07$} \\

\end{longtable}
\endgroup

\end{document}